\documentclass[10pt,twocolumn,letterpaper]{article}

\usepackage{iccv}
\usepackage{times}
\usepackage{epsfig}
\usepackage{graphicx}
\usepackage{amsmath}
\usepackage{amssymb}
\usepackage{booktabs}
\usepackage{multirow}
\usepackage{adjustbox}
\usepackage{nicefrac}

\usepackage[pagebackref=true,breaklinks=true,letterpaper=true,colorlinks,bookmarks=false]{hyperref}

\iccvfinalcopy 

\def\iccvPaperID{7731} 

\ificcvfinal\fi

\begin{document}

\title{RGBD-Net: Predicting color and depth images for novel views synthesis}

\author{
Phong Nguyen \\
University of Oulu
\and
Animesh Karnewar \\
TomTom
\and
Lam Huynh \\
University of Oulu
\and
Esa Rahtu \\
Tampere University
\and
Jiri Matas \\
Czech Technical University in Prague
\and
Janne Heikkil\"a \\
University of Oulu
}

\maketitle
\ificcvfinal\thispagestyle{empty}\fi

\begin{abstract}
We propose a new cascaded architecture for novel view synthesis, called RGBD-Net, which consists of two core components: a hierarchical depth regression network and a depth-aware generator network. The former one predicts depth maps of the target views by using adaptive depth scaling, while the latter one leverages the predicted depths and renders spatially and temporally consistent target images. In the experimental evaluation on standard datasets, RGBD-Net not only outperforms the state-of-the-art by a clear margin, but it also generalizes well to new scenes without per-scene optimization. Moreover, we show that RGBD-Net can be optionally trained without depth supervision while still retaining high-quality rendering. Thanks to the depth regression network, RGBD-Net can be also used for creating dense 3D point clouds that are more accurate than those produced by some state-of-the-art multi-view stereo methods.

\end{abstract}

\begin{figure*}[t]
\begin{center}
   \includegraphics[width=\linewidth]{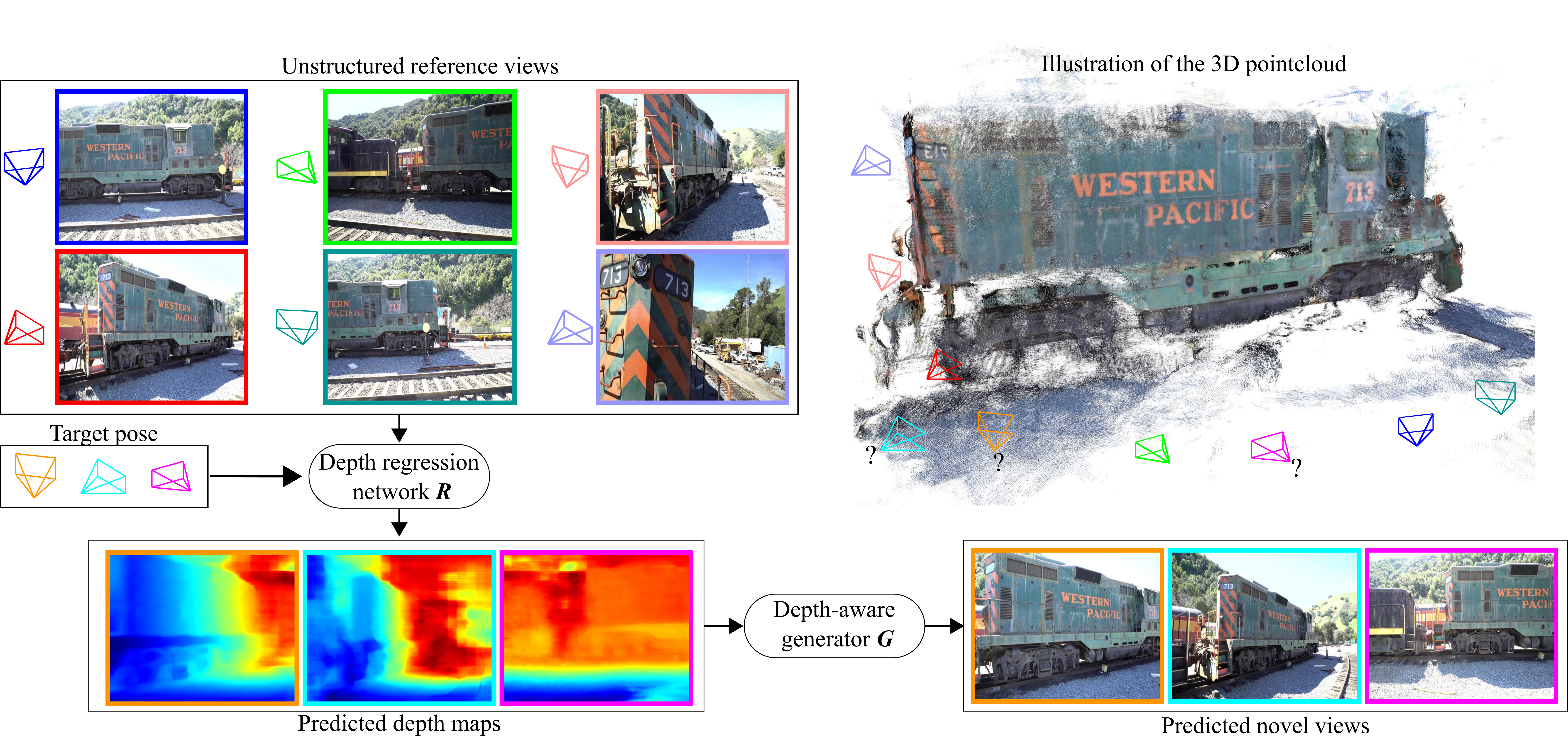}
\end{center}
   \caption{\textbf{RGBD-Net} is a novel view synthesis method that estimates both the depth maps and color images of novel views from significantly different viewpoints.
   From the estimates, we reconstructed a point-cloud of the Train scene of the Tanks and Temples dataset \cite{TankAndTemples}.
   Note: We trained {RGBD-Net} on multi-view stereo datasets \cite{DTU,yao2020blendedmvs} and the Tanks and Temples dataset is used only in evaluation.}
\label{fig_intro}
\end{figure*}

\section{Introduction}
Novel View Synthesis (NVS), also called Image-Based Rendering (IBR), is a long-standing problem that has applications in free-viewpoint video, telepresence, and mixed reality \cite{VR}.
NVS is a problem where visual content is captured from one or several reference views and synthesized for an unseen target view. 
The problem is challenging since mapping between views depends on the 3D geometry of the scene and the camera poses between the views. Moreover, NVS requires not only  propagation of information between views but also hallucination of details in the target view that are not visible in reference image due to occlusions or limited field of view.

Early NVS methods produced target views by interpolating in ray \cite{intro_ray} or pixel space \cite{intro_pixel}. 
They were followed by works that leveraged certain geometric constraints such as epipolar consistency \cite{intro_warp} for depth-aware warping of the input views. These interpolation based methods suffered from artifacts arising from occlusions and inaccurate geometry.
Later works tried to patch the artifacts by propagating depth values to similar pixels \cite{intro_similardepth} or by soft 3D reconstruction \cite{intro_soft3d}. 
However, these approaches cannot leverage depth to refine the synthesized images or deal with the unavoidable issues of temporal inconsistency.

More recently, functional representation methods \cite{neuralrep3,Nerf,SparseVoxelFields} have employed deep neural networks to learn implicit scene representations from a large set of observations of a specific scene. 
Although these methods produce impressive novel views, dedicated per-scene training is required to apply the representation to a new scene. 
Another research direction \cite{StereoMag,LLFF,EVS} uses a small number of observations at each training step.  
While the quality of the generated novel views of these methods is worse than those produced by the function-based representation methods, they generalize to unseen data without fine-tuning or retraining.


In this paper, we adopt the best of both approaches and develop a method that renders high-quality novel views from an unstructured set of reference images, without needing per-scene training. We propose a new method called RGBD-Net that produces both color (RGB) and depth (D) images of the unseen target view. As illustrated in Fig.~\ref{fig_intro}, RGBD-Net includes two main modules: a depth regression network $R$ and a depth-aware generator network $G$. The first module estimates the target view depth map and the second module produces photo-realistic novel views using the regressed depth maps. Our experiments show that it generalizes well to arbitrary scenes without the need for per-scene optimization.
To summarize, the main contributions of our work are:
\begin{itemize}
    \item Adaptive depth scaling that enables producing photo-realistic novel views with and without per-scene optimization.
    \item A spatial-temporal module that allows RGBD-Net to produce a smooth sequence of rendered novel views along a continuous camera path.
    \item State-of-the-art results in novel view synthesis on three challenging large-scaled 3D datasets~\cite{DTU,yao2020blendedmvs,TankAndTemples}.
\end{itemize}
Source code and neural network models will be made publicly available upon publication of the paper.

\section{Related Work}
In the following, we discuss different methods of novel view synthesis. For an extensive discussion of view synthesis, we would like to refer to Tewari et al. \cite{report}.

\noindent{\textbf{\textit{Novel view synthesis.}}} Early works on view synthesis with deep learning often use a Plane Sweep Volume (PSV) \cite{PSV} for novel view synthesis.
Each input image is projected onto successive virtual planes of the target camera to form a PSV. Kalantari et al. \cite{Kalantari} calculates the mean and standard deviation per plane of the PSV to estimate the disparity map and render the target view. {RGBD-Net} proposes a hierarchical depth regression network that predicts the depth map of the novel view in a coarse-to-fine manner. Moreover, our work focuses on solving the problem of view synthesis using unstructured inputs which poses a more demanding challenge than using grid-sampled views captured by the light-field camera. Extreme View Synthesis (EVS) \cite{EVS} builds upon DeepMVS \cite{DeepMVS} to estimate a depth probability volume for each input view that is then warped and fused into the target view. Rather than estimating the depth maps of the source images, we train {RGBD-Net} to predict the depth map at the target view directly and then refine the warped novel images using a depth-aware generator network.

Perhaps, the closest work to {RGBD-Net} is the recently published Free View Synthesis (FVS) by Riegler et al. \cite{FVS}. In this work, they use a structure-from-motion method \cite{colmap} to reconstruct a 3D mesh of the scene for creating an incomplete depth map for the target view. They also propose a recurrent blending network to refine the warped novel views. {RGBD-Net} estimates a complete depth map to refine the warped novel image.

\noindent{\textbf{\textit{Neural scene representations.}}} 
Recent geometric deep learning methods learn to deal with 3D scenes using various types of 3D representations such as multi-layered representation~\cite{StereoMag,MPI2,MPI3,deepview}, voxel-grids~\cite{neuralVolumes,Deepvoxels}, meshes~\cite{neuralrep4}, point-clouds~\cite{synsin,NPBG}, and function-based~\cite{neuralrep1, SparseVoxelFields, crowdsampling, NPBG, Nerf, xfield, neuralrep3}. 

A significant number of works \cite{StereoMag,MPI2,MPI3,deepview} on view synthesis represent the 3D scene by Multiple Plane Images (MPIs). Each MPI includes multiple RGB-$\alpha$ planes, where each plane is related to a certain depth. The target view is generated by using alpha composition \cite{alpha} in the back-to-front order. Zhou et al. \cite{StereoMag} introduce a deep convolutional neural network to predict MPIs that reconstruct the target views for the stereo magnification task. 
Later work by Flynn et al. \cite{deepview} considerably improves the quality of synthesized images in the light-field setups. They propose a novel network with a regularized gradient descent method to refine the generated images gradually.
Local Light Field Fusion (LLFF) \cite{LLFF} introduces a practical high-fidelity view synthesis model that blends neighboring MPIs to the target view. The input to the MPI-based methods is also PSVs. However, those PSVs are constructed on a fixed range of depth values. The proposed RGBD-Net builds multi-scale PSVs which use adaptive sampled depth planes. 

Grid-based representations are similar to the MPI representation, but are based on a dense uniform grid of voxels. This representation have been used as the basis for neural rendering techniques to model object appearance~\cite{neuralVolumes,Deepvoxels}. Sitzmann et al.~\cite{Deepvoxels} learns a persistent 3D feature volume for view synthesis and employs learned ray-marching. Neural Volumes~\cite{neuralVolumes} is an approach for learning dynamic volumetric representations of multi-view data. The main limitation of grid-based methods is the required cubic memory footprint. The sparser the scene, the more voxels are empty, which wastes model capacity and limits output resolution. We propose a memory efficient multi-scale PSV representation which assigns dynamic depth planes per-pixel.

Recent works~\cite{synsin,NPBG,Pulsar} on view synthesis have also employed the point-based representation to model 3D scene appearance. A drawback of the point-based representation is that there might be holes between points after projection to the screen space. Aliev et al. \cite{NPBG} trains a neural network to learn feature vectors that describe 3D points in a scene. These learned features are then projected onto the target view and fed to a rendering network to produce the final novel image. Wiles et al.~\cite{synsin} lifts per-pixel features from a source image onto a 3D pointcloud that can be explicitly projected to the target view using a U-Net model. However, this method suffers from temporal instabilities between generated novel views of a smooth camera path. We propose to model the spatial-temporal relations between queried target poses to render a smooth sequence of novel views without per-scene optimization~\cite{stable,Nerf}.  

The current state-of-the-art method Neural Radiance Fields (NeRF) by Mildenhall et al. \cite{Nerf} represents the plenoptic function by a multi-layer perceptron that can be queried using classical volume rendering to produce novel images. NeRF has to be evaluated at a large number of sample points along each camera ray. This makes rendering a full image with NeRF extremely slow. Despite the high quality of the synthesized novel images, NeRF also requires per-scene training. Recent volumetric approaches \cite{yu2020pixelnerf,wang2021ibrnet,hybridnerf,trevithick2020grf} address the generalization issue of NeRF by incorporating a latent vector extracted from reference views. These methods show generalization on selected testing scenes, but they share the slow rendering property of NeRF \cite{Nerf}. We propose a novel adaptive depth scaling to produce up-to-scale depth maps on various types of 3D scenes. Thus, RGBD-Net achieves good performance on a testing set that is separate from the training set and completely new scenes outside those datasets. Since our method is fully convolutional, RGBD-Net also achieves faster rendering than NeRF and its variants.

\begin{figure*}[ht]
\begin{center}
   \includegraphics[width=1.0\linewidth]{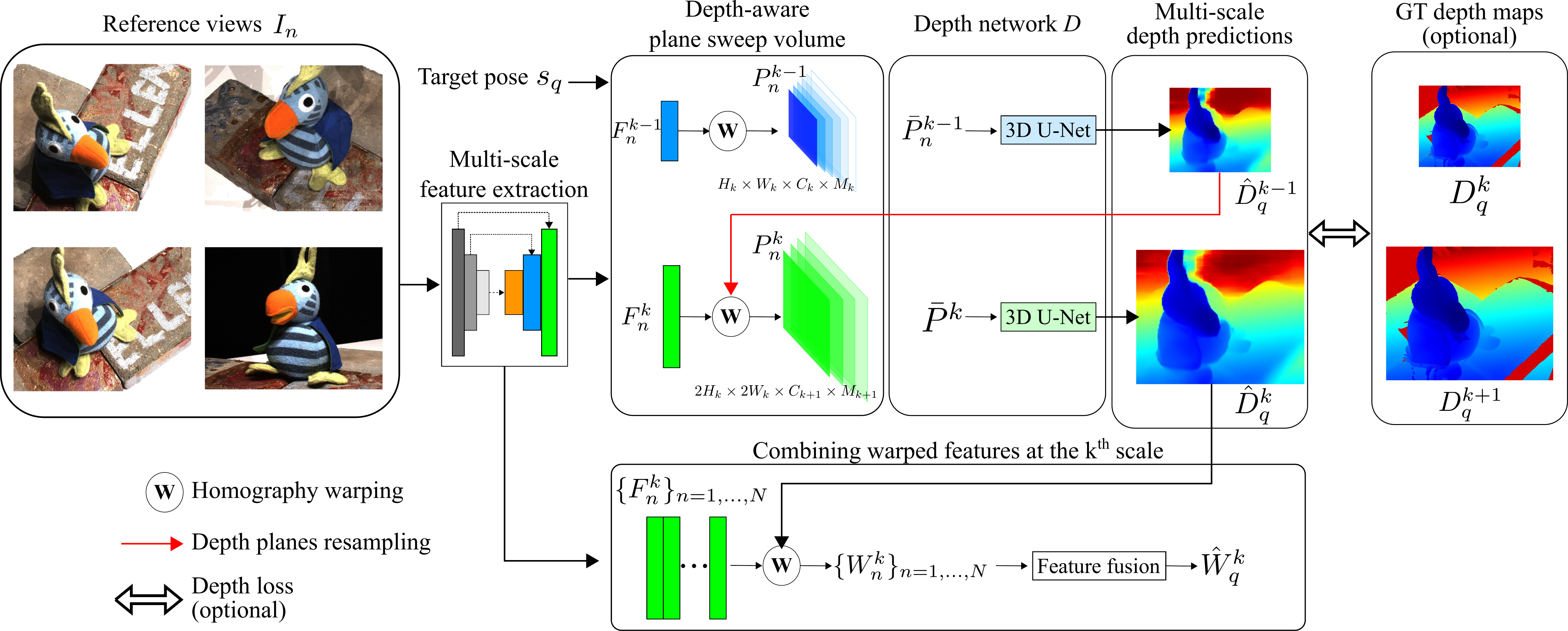}
\end{center}
   \caption{The architecture for estimating the depth map of the novel view.
   Note that, the method does not require explicit ground-truth depth loss for training and is able to predict plausible depth maps of the target views using only the RGB image loss. 
   The predicted depth map is used to warp the features, extracted from the reference views, onto the target view and then subsequently passed to the $G$ network. }
\label{fig_depthNet}
\end{figure*}

\section{Proposed method}
This section describes in detail the architecture of {RGBD-Net}, which comprises of two modules: a hierarchical depth regression network $R$ (Section \ref{depth_net}) that estimates the depth map of the novel view, and a depth-aware refinement network $G$ (Section \ref{gen_net}) that enhances the warped images to produce the final target image. Last, we discuss the loss functions used to train the model in Section~\ref{loss}.

\subsection{Depth regression network $R$} \label{depth_net}
We first describe the pipeline (see Fig. \ref{fig_depthNet}) for estimating the depth map $\hat{D}_q$ of the target view $s_q$ from a set of unstructured input images and their poses $\{I_n,s_n\}_{n=1,...,N}$. Each reference view $I_n$ is first fed to the Feature Pyramid Network \cite{FPN} to extract $K$ multi-scale features $F^k_n$ \cite{MSGGAN}. We then apply homography warping to each feature map of $F^k_n$ to construct a PSV $P^k_n$ of the target view $s_q$ with a set of $M_k$ hypothesis depth planes. A mean PSV $\bar{P}^k=\sum_{n=1}^{N} P^k_n / N$ is fed to a 3D U-Net to estimate a coarse novel depth map $\hat{D}^k_q$. Inspired by the recent work on multi-view stereo~\cite{MVSNet,CasMVSNet}, we estimate the depth map of the novel views in a coarse-to-fine manner. We first utilize the depth plane resampling technique from the MVS literature to efficiently sample $M_{k+1}$ depth planes using the predicted coarse novel depth map $\hat{D}^k_q$.

\noindent \textbf{Depth plane resampling.} The depth planes $d_i^1$ at the scale $k=1$ are sampled from the initial depth range as follows:
\begin{equation}
\label{eqn_overall}
d_i^1 = d^1_{\min} + i\Delta_1, \quad i = 1,..,M_1
\end{equation}
where $d^1_{\min}$ and $\Delta_1$ are the minimum depth value and depth interval, respectively. At the later stages ($k>1$), the adjusted depth ranges \textit{per pixel} are selected such that, their centers lie at the estimated depth values obtained from the depth map of the previous stage. We then rewrite equation (\ref{eqn_overall}) to define the sampled depth plane $d_i^k(p)$ for a pixel $p$ as follows:
\begin{equation}
\label{eqn_rewrite}
d_i^k(p) = d^k_{\min}(p) + i\Delta_k, \quad i = 1,..,M_k
\end{equation}
\begin{equation}
\label{eqn_dMIN}
d^k_{\min}(p) = \hat{D}^{k-1}_q(p) - \frac{M_k\Delta_k}{2}
\end{equation}
where $\hat{D}^{k-1}_q(p)$ is the predicted depth value of the pixel $p$ from the last stage. Instead of having a constant minimum depth value $d^k_{\min}$, we leverage $\hat{D}^{k-1}_q$ to obtain adaptive $d^k_{\min}(p)$ for each pixel $p$. The adaptive $d^k_{\min}(p)$ narrows the sampled depth ranges and allows RGBD-Net to produce more accurate depth maps. The width and height of the PSVs are doubled when $k$ is increased by one (see Fig.~\ref{fig_depthNet}). Therefore, we set $M_k = M_{k-1}/2$ and $\Delta_k = \Delta_{k-1}/2$ to narrow the depth range of the subsequent stage.

\noindent \textbf{Adaptive depth scaling} RGBD-Net focuses on solving the generalization problem of view synthesis. To address this issue, we propose a scaling method to handle 3D scenes in various depth ranges. In practice, some depth ranges are from 0.1 to 1 or from 10 to 100. Note that, these numbers are not the absolute distances in some known units. Thus, we transform those depth ranges roughly to the same scale. Let $d_{\min}$ and $d_{\max}$ be the minimum and maximum depth values of an arbitrary 3D scene. We define a scaling factor $f=C/d_{\min}$ where $C$ is a constant value. 
The minimum depth value of the depth plane resampling is then scaled so that $d^1_{min}=C$.
Based on the per-scene scaling factor $f$, we can obtain the scaled depth interval $\Delta_1 = (f d_{\max} - C)/M_1$. In all experiments, we use the same the number of hypothesis depth plane $M_1$ to save GPU memory.

If the ground-truth depth is available, we can train the $R$ network with depth supervision. The ground-truth depth is scaled using the same scaling factor $f$. This method encourages depth robustness on various depth scales without per-scene optimization.
If the ground-truth depth is not available, we use COLMAP~\cite{colmap} to perform sparse reconstruction and get the depth range of such testing scenes. We then scale the obtained depth range using the similar technique. The novel view evaluation in the Section~\ref{experiments} validates the generalization ability of RGBD-Net.

\subsection{Depth-aware refinement network $G$}\label{gen_net}

\noindent \textbf{Feature fusion.}\label{fusion}
We use differentiable bilinear interpolation from Jaderberg et al. \cite{STN} to map the learned features $F^k_n$ of the Feature Pyramid Network to obtain the warped feature $W^k_n$ using the previously regressed depth map. 
We then combine the set $\{W^k_n\}_{n=1,...,N}$ to obtain the unified warped feature $\hat{W^k_q}$ of the target pose $s_q$.
We rely on the predicted depth map to calculate the 3D coordinates of every pixel $p$ of the novel view. 
Using the known reference and target pose, we back-project that 3D point to the camera space of each reference view. 
The unified warped feature $\hat{W}^k_q$ of the target pose is obtained as follows:
\begin{equation}
    \alpha_n^p = (1/z_n^p)/\sum_N 1/z_n^p
\end{equation}
\begin{equation}
    \hat{W}^k_q = \sum^N \alpha_n W^k_n
\end{equation}
where $z_n^p$ and $\alpha_n^p$ are the z-coordinate and the blending weight of the pixel $p$ of the $n^{th}$ reference view respectively. If the pixel $p$ is not visible in all reference views then its weight is zero. We utilize the inverse of the z-coordinate to reduce the impact of far reference views from the target pose. Therefore, the blending weights $\alpha_n$ rely on the predicted depth map to assign weights between reference views. As the depth network $R$ gets better at predicting depth maps, so does the feature fusion method.

\begin{figure}[t]
\begin{center}
   \includegraphics[width=0.99\linewidth]{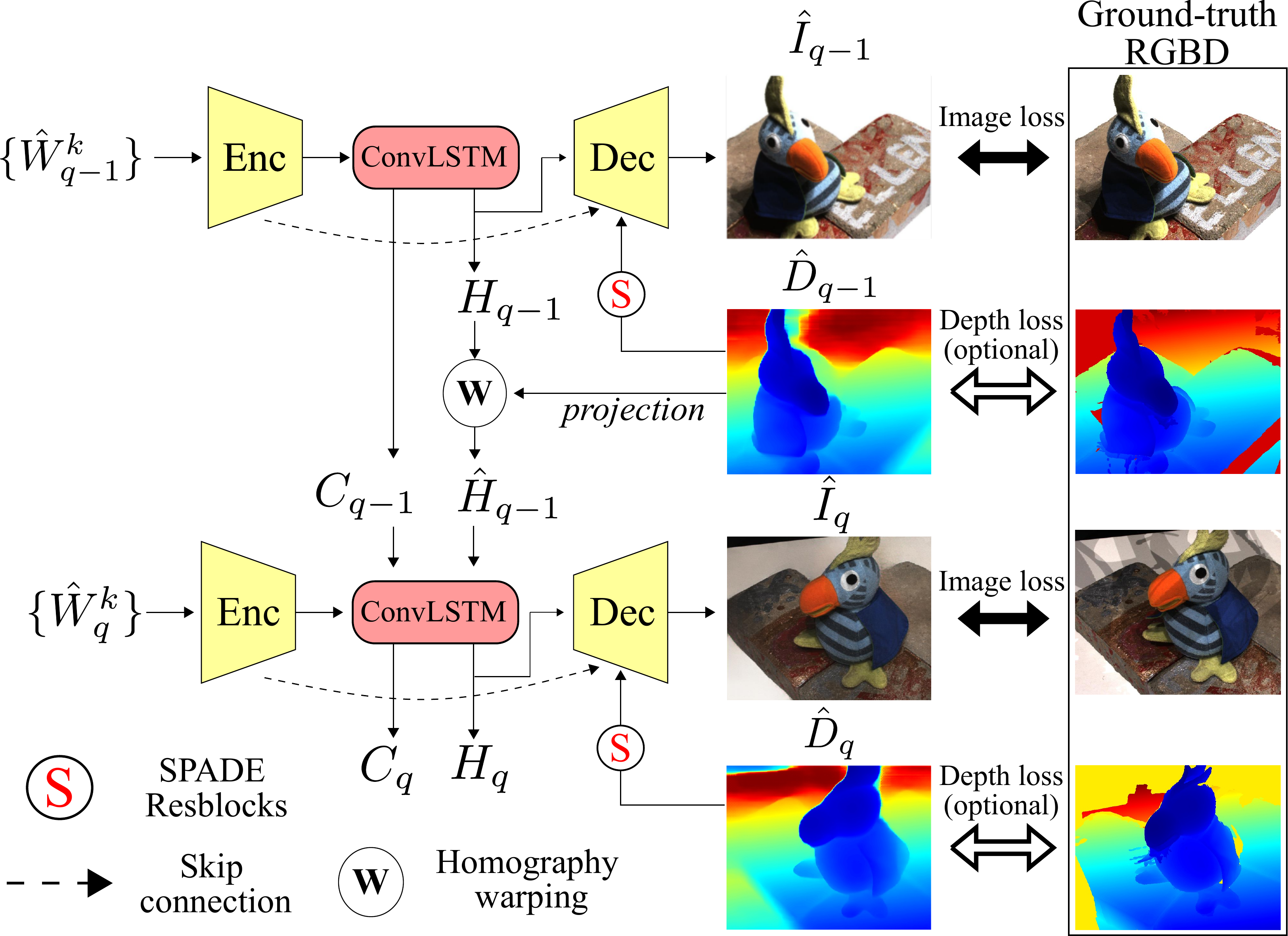}
\end{center}
   \caption{We encourage a spatial-temporal consistency by training the depth-aware refinement network $G$ to render $Q$ nearby sampled novel views. Each novel view $\hat{I}_q$ is predicted from a set a set of $K$ multi-scale warped features $\{\hat{W}^k_q\}_{k=1,...,K}$.}
\label{fig_genNet}
\end{figure}

\noindent \textbf{Depth-aware synthesis network.}\label{depth-aware} 
As illustrated in Fig.~\ref{fig_genNet}, the proposed depth-aware refinement network $G$ predicts the novel view $\hat{I}_q$ of the target pose $s_q$ using a set of $K$ multi-scale features $\{\hat{W}^k_q\}$. We provide details of the 2D U-Net structure in the supplementary material.

To further enhance the overall quality of the predicted novel image, we leverage the complete predicted depth maps produced by the depth regression network $R$. Recent successes on conditional image synthesis \cite{spade,sean,styleGAN2} have shown that we can generate photo-realistic images conditioned on certain image data such as an image from another domain or a semantic segmentation map. We observe that the proposed $R$ network  produces depth maps with sharp edges. Therefore, we exploit this to guide the image synthesizing model to produce sharp novel images. 

The predicted depth maps $\hat{D}_q$ are fed to the Spatially-Adaptive Denormalization (SPADE) Resblocks \cite{spade} to progressively predict the novel views in a coarse-to-fine manner. More specifically, each SPADE Resblock regularizes the decoder's learned features based on the depth predictions. This transformation ensures that the predicted novel image has similar sharp edges as the depth map produced by the $R$ network. Besides, utilizing the depth maps in the SPADE blocks provides a valid inductive bias to the refinement network for synthesizing the novel views even without explicitly learning the depth maps with ground truth.

\noindent \textbf{Spatial-temporal consistency.}\label{consistency} RGBD-Net uses a set of sparse reference views to synthesize a novel view. Therefore, when generating videos along smooth camera paths, it is potentially subject to temporally inconsistent predictions and flickering artifacts due to the independent rendering at each new viewpoint.

To address the above issue, we include a ConvLSTM~\cite{ConvLSTM} cell to model the spatial-temporal relations between $Q$ randomly sampled novel views.
Conventional LSTM utilizes the previous hidden state $H_{q-1}$ to produce the current hidden state $H_q$. In view synthesis, there are certain changes between viewpoints of the $Q$ target views. Hence, we reflect these viewpoint changes by warping~\cite{STN} $H_{q-1}$ using the previously predicted depth map $\hat{D}_{q-1}$. The warped hidden state $\hat{H}_{q-1}$ represents the encoded image feature of the previous novel view being warped to the current target pose $s_q$. Let $O_q$ denote the output of the encoder, we obtain the novel view $\hat{I}_q$ using ConvLSTM~\cite{ConvLSTM} as follows:
\begin{equation}
    \hat{H}_{q-1} = \text{warping}(H_{q-1},\hat{D}_{q-1})
\end{equation}
\begin{equation}
\label{eqn_warping}
    C_q,H_q = \text{ConvLSTM}(O_q,C_{q-1},\hat{H}_{q-1})
\end{equation}
\begin{equation}
    \hat{I}_q = \text{Dec}(H_q,S_q)
\end{equation}
where $S_q$ is the skip connections from the encoder to the decoder. Instead of using the final output $O_q$ of the encoder to render the novel view, we use the output hidden state $H_{q}$ of the ConvLSTM as the decoder's input to render the novel view $\hat{I}_q$. As can be seen in (\ref{eqn_warping}), the learned ConvLSTM cell aggregates the current and prior encoded visual features of $O_q$ and  $\hat{H}_{q-1}$ to eliminate the temporal inconsistency between adjacent target poses.

\subsection{Training}\label{loss}
\noindent\textbf{Learning objective.} We trained the proposed method with an L1 image loss $\mathcal{L}_{l1}$, perceptual loss $\mathcal{L}_{p}$ \cite{vggLoss} and hinge GAN loss $\mathcal{L}_{G}$ \cite{GAN} between the generated and ground-truth novel image. If the ground-truth depth map is available we can also use the scaled depth loss $\mathcal{L}_{d}$~\cite{depthLosses}.
The total loss is then $\mathcal{L}_{total} = \lambda_{l1}\mathcal{L}_{l1} + \lambda_p\mathcal{L}_{p} + \lambda_{G}\mathcal{L}_{G} + \lambda_{d}\mathcal{L}_{d}$. Note that our method does not strictly need the depth loss $\mathcal{L}_{d}$, which enables training on datasets that do not have ground-truth depth maps. \\
\noindent\textbf{Implementation details.} The models were trained with the Adam optimizer using a 0.004 learning rate for the discriminator, 0.001 for both the depth regression $R$ and refinement generator $G$ and momentum parameters (0, 0.9). $\lambda_{l1} = 1, \lambda_p = 10, \lambda_{G} = 1,\lambda_{d} = 1, K = 3, N = 7, C = 100, Q = 3, W = 640, H = 512$. We implemented {RGBD-Net} in PyTorch \cite{Pytorch}, and training took 2-3 days on 4 Tesla V100 GPUs.

\section{Experiments} \label{experiments} 

\begin{table*}[ht]
\resizebox{\textwidth}{!}{
\centering
\begin{tabular}{@{}lcccccccccccc@{}}
\toprule
\multicolumn{1}{c}{\multirow{2}{*}{Method}} &
  \multicolumn{3}{c}{Tank\&Temples~\cite{TankAndTemples}} &
  \multicolumn{3}{c}{DTU~\cite{DTU}} &
  \multicolumn{3}{c}{BlendedMVS~\cite{yao2020blendedmvs}} &
  \multicolumn{3}{c}{Real Forward-Facing~\cite{LLFF}} \\ \cmidrule(l){2-13} 
\multicolumn{1}{c}{} & LPIPS$\downarrow$ & SSIM$\uparrow$  & PSNR$\uparrow$  & LPIPS$\downarrow$ & SSIM$\uparrow$  & PSNR$\uparrow$  & LPIPS$\downarrow$ & SSIM$\uparrow$  & PSNR$\uparrow$  & LPIPS$\downarrow$ & SSIM$\uparrow$  & PSNR$\uparrow$  \\ \midrule
pixelNeRF~\cite{yu2020pixelnerf}            & 0.65  & 0.496 & 12.25 & 0.54  & 0.857 & 19.25 & 0.48  & 0.724 & 16.28 & 0.44  & 0.638 & 20.15 \\
LLFF~\cite{LLFF}                 & 0.61  & 0.524 & 13.25 & 0.51  & 0.872 & 21.25 & 0.41  & 0.794 & 17.28 & 0.22  & 0.798 & 24.23 \\
FVS~\cite{FVS}                 & 0.18  & 0.868 & 20.26 & 0.25  & 0.972 & 26.96 & 0.25  & 0.815 & 22.94 & 0.21  & 0.815 & 25.67 \\
$\text{RGBD-Net}^*$             & 0.17  & 0.884 & 20.35 & 0.21   & 0.980 & 31.69 & 0.21  & 0.838 & 23.52 & 0.20  & 0.832 & 26.51 \\
RGBD-Net &
  \textbf{0.16} &
  \textbf{0.892} &
  \textbf{21.28} &
  \textbf{0.19} &
  \textbf{0.985} &
  \textbf{32.65} &
  \textbf{0.18} &
  \textbf{0.859} &
  \textbf{25.13} &
  \textbf{0.18} &
  \textbf{0.854} &
  \textbf{26.92} \\ \midrule
$\text{NPBG}^{\dagger}$~\cite{NPBG}               & 0.24  & 0.821 & 19.46 & 0.36  & 0.942 & 24.78 & 0.36  & 0.801 & 20.18 & - & - & - \\
$\text{NeRF++}^{\dagger}$~\cite{nerfplus}               & 0.14  & 0.952 & 25.69 & 0.14  & 0.991 & 35.28 & 0.15  & 0.913 & 25.58 & 0.18  & 0.867 & 26.55 \\
$\text{RGBD-Net}^{\dagger}$  &
  \textbf{0.12} &
  \textbf{0.986} &
  \textbf{26.35} &
  \textbf{0.11} &
  \textbf{0.997} &
  \textbf{36.69} &
  \textbf{0.09} &
  \textbf{0.935} &
  \textbf{29.52} &
  \textbf{0.16} &
  \textbf{0.882} &
  \textbf{27.45} \\ \bottomrule
\end{tabular}
}
\caption{Quantitative comparison on large-scale dataset of synthetic and real images.  
For all datasets, the metrics (average over all target views) are reported. The RGBD-Net is trained with combined loss $\mathcal{L}_{total}$, achieving best results without per-scene optimization.
Performance without the depth loss $\mathcal{L}_{d}$, denoted RGBD-Net$^*$, is competitive.
Methods with a $\dagger$ symbol are optimized per-scene.
Finetuning scene-specific $\text{RGBD-Net}^{\dagger}$ achieves state-of-the-art results on view synthesis.}
\label{table_quant_NVS}
\end{table*}

\begin{figure*}
\begin{center}
  \includegraphics[width=\linewidth]{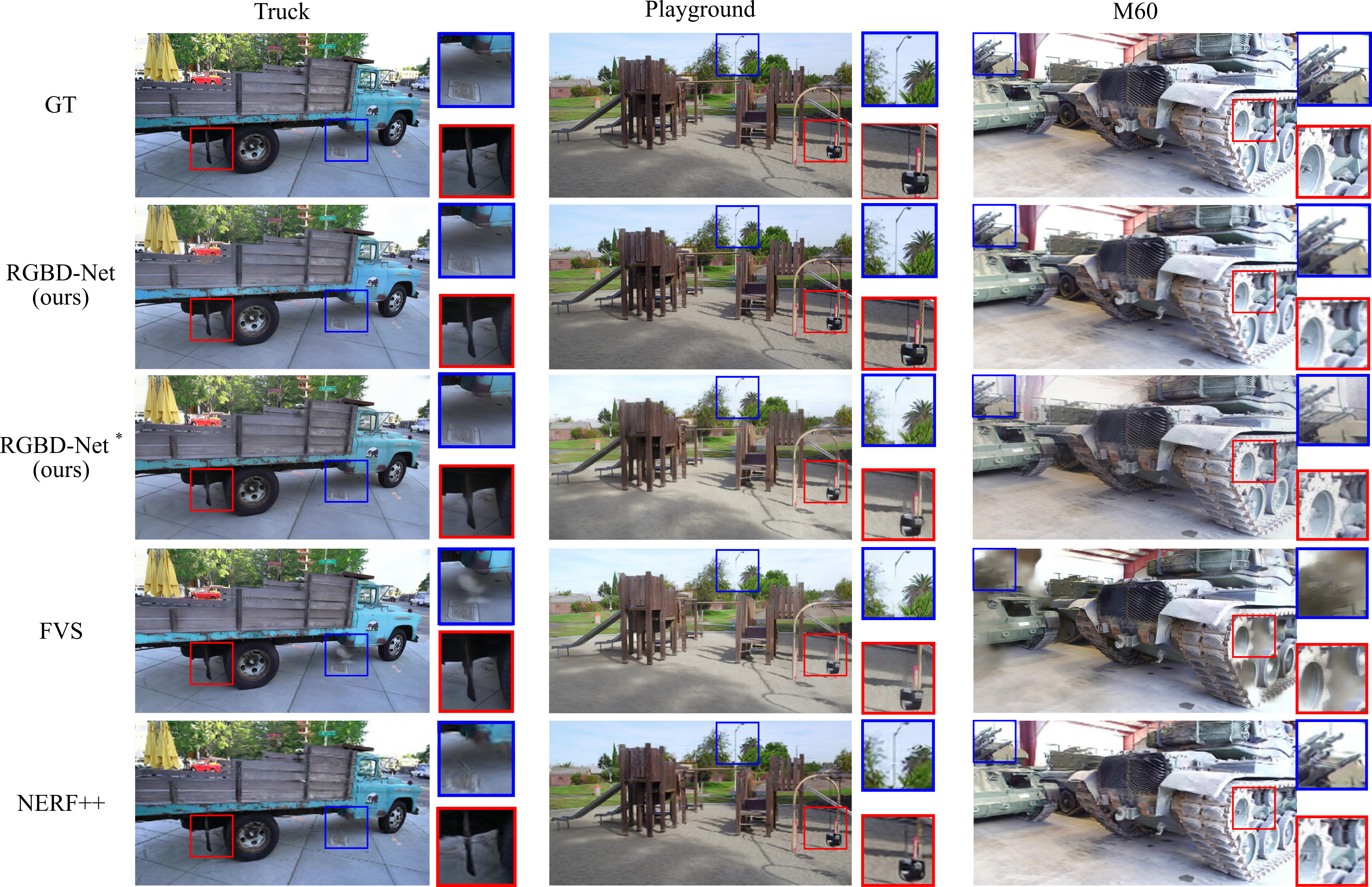}
\end{center}
  \caption{Exampled of generated novel views by RGBD-Net and state-of-the-art methods for three scenes from the Tanks and Temples (T$\&$T) \cite{TankAndTemples} dataset. We train RGBD-Net on the DTU~\cite{DTU} dataset and test it on T$\&$T to evaluate the generalization ability.
  }
\label{fig_qualNVS}
\end{figure*}

\noindent\textbf{View selections.} We follow the view selection method of Riegler et al. \cite{FVS} to select the top 10 closest source images to each target image. During training, we randomly select $N$ source images among the 10 closest views as inputs to our method. At each training step, we sample $N$ uniformly at random from $[1,N]$.
For each target pose, we randomly select $Q-1$ nearby target poses. We train RGBD-Net to produce $Q$ target poses in each forward pass to encourage the temporal consistency.
\\
\noindent\textbf{Datasets.} We train {RGBD-Net} using the DTU \cite{DTU} and BlendedMVS \cite{yao2020blendedmvs} datasets. DTU is an MVS dataset consisting of more than 100 scenes scanned in 7 different lighting conditions at 49 positions. From 49 camera poses, we selected 10 as targets for view synthesis and used the rest for source image selection.
BlendedMVS \cite{yao2020blendedmvs} is another large-scale MVS dataset which contains high quality rendered and real images with realistic ambient lighting. We first train {RGBD-Net} on the DTU training set and then fine-tune it on the BlendedMVS training set. We found that this yields better performance than training from scratch. We evaluate the performance of our model using the DTU and BlendedMVS testing sets. To test the generalization capability of our method, we  test it on the intermediate set of the Tanks and Temples \cite{TankAndTemples} dataset. We use the same dedicated camera path of the testing scenes to evaluate RGBD-Net against recently proposed \cite{FVS} and other NVS methods. Finally, we evaluate RGBD-Net on real forward-facing scenes~\cite{LLFF}. Each scene includes 12 to 62 images and $1/8$ of these images are held out for testing.

\begin{figure}[t]
\begin{center}
  \includegraphics[width=\linewidth]{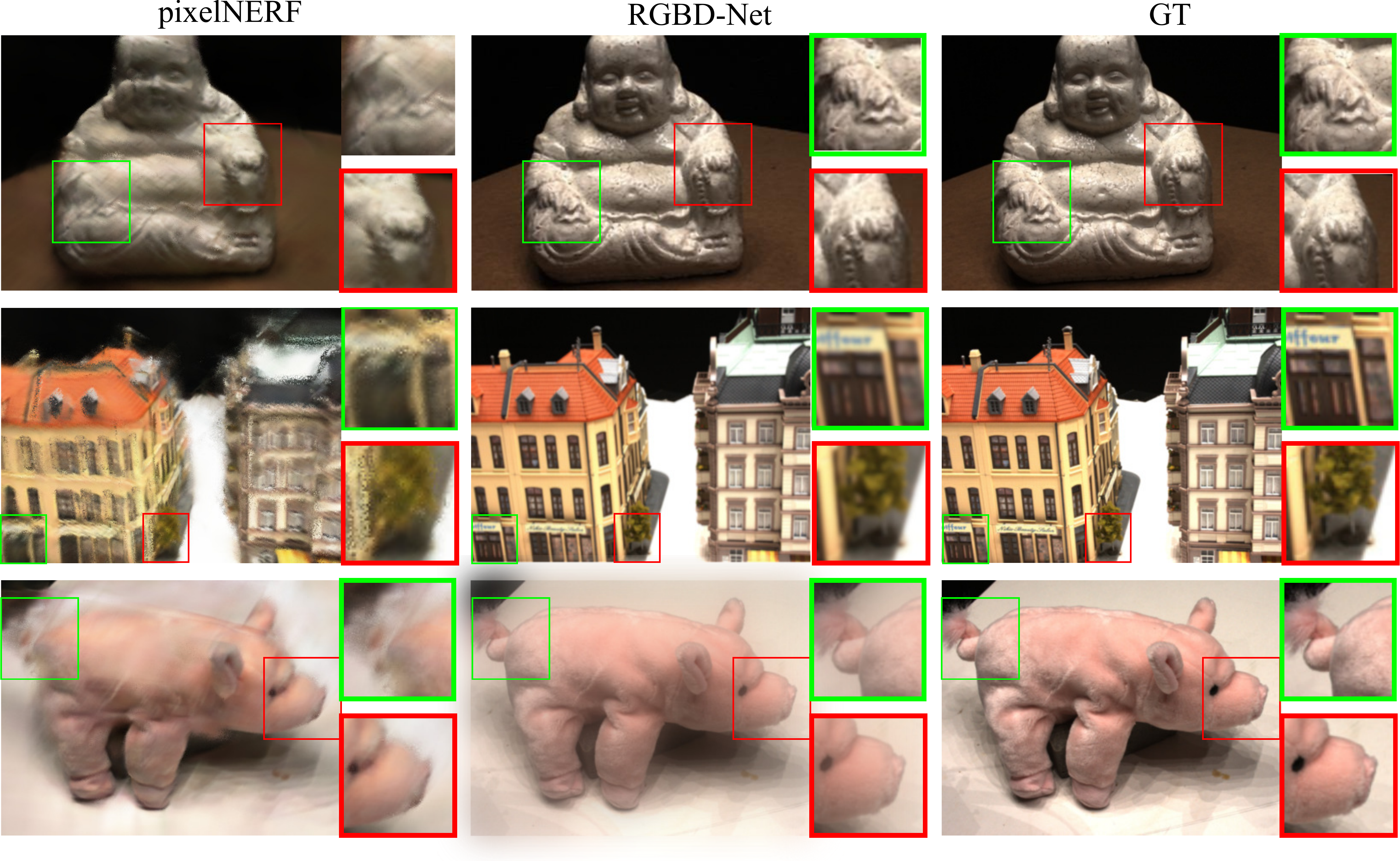}
\end{center}
  \caption{ Novel views generated by RGBD-Net and recently proposed pixelNeRF~\cite{yu2020pixelnerf} on for three test scenes from the DTU~\cite{DTU}. 
  }
\label{fig_pixelNERF}
\end{figure}

\noindent\textbf{Baselines.} In this evaluation, we compare our approach to recently proposed methods on novel view synthesis. To evaluate the generalization ability of RGBD-Net, we compare with pixelNeRF~\cite{yu2020pixelnerf} which is a variant of NeRF that generalizes on multiple 3D scenes. We use the provided pretrained pixelNeRF on the DTU dataset~\cite{DTU} and compare with ours. Furthermore, we compare RGBD-Net with other view synthesis methods: LLFF \cite{LLFF} and FVS \cite{FVS}. We use the provided pre-trained models of these methods to produce novel images and compare them with ours. 

Among the current state-of-the-art neural rendering methods, NeRF++ \cite{nerfplus} and Neural Point-based Graphics (NPBG) \cite{NPBG} require retraining on the test scenes, whereas our method does not need any adaptation or fine-tuning on new scenes. We use their public training code to train their models for each test scene. Instead of vanilla NeRF~\cite{Nerf}, we choose to compare with NeRF++~\cite{nerfplus} because the method performs better on the large-scale, unbounded 3D scenes.
\\
\noindent\textbf{Metrics.} We report the PSNR, SSIM, and perceptual similarity (LPIPS) \cite{LPIPS} of view synthesis between {RGBD-Net} and other state-of-the-art methods.
\\
\noindent\textbf{Results.} We summarize the quantitative and qualitative results in Table \ref{table_quant_NVS} and Fig.~\ref{fig_qualNVS}. The model {$\text{RGBD-Net}^*$}, which is trained without the ground truth depth loss $\mathcal{L}_{d}$, shows almost similar performance to the full model, while still being better than other baselines. We also observe no significant differences between the predicted novel views produced by RGBD-Net when trained with or without depth supervision. The goal of view synthesis is to produce faithful novel views and for that purpose, we do not strictly need the predicted depth map at the target view to be perfectly accurate. 


We first evaluate RGBD-Net against the current top-performing image-based rendering methods~\cite{LLFF,FVS,yu2020pixelnerf} which do not require per-scene optimization. As can be seen in Fig.~\ref{fig_qualNVS}, FVS~\cite{FVS} struggles to recover clean and accurate boundary (ghosting artifacts in M60 and Truck), and fails to capture thin structures (lamp post on Playground). Although this methods has been trained on the the training set of Tanks and Temples dataset~\cite{TankAndTemples}, our base model RGBD-Net can render more realistic novel views compared those produced by FVS~\cite{FVS}.
We also test the generalization of  RGBD-Net on the Real Forward-facing~\cite{LLFF} dataset.
LLFF~\cite{LLFF} perform reasonably well on this dataset because the method is based on the MPI representation and assumes that camera poses lie on the same plane.
The reference views of RGBD-Net does not need to follow that assumption. 
Experimental results show that our method performs substantially better than LLFF~\cite{LLFF} on the  real-world scenes.

PixelNeRF~\cite{yu2020pixelnerf} is a recent approach to extend NeRF~\cite{Nerf} for the generalization. Quantitative results show that this method does not perform well on the Tanks and Temples dataset~\cite{TankAndTemples}. To fairly evaluate the generalization ability of RGBD-Net against pixelNeRF~\cite{yu2020pixelnerf}, we evaluate them on the testing set of the DTU dataset~\cite{DTU}. In Fig.~\ref{fig_pixelNERF}, we notice the lack of fine details in the novel views produced by pixelNeRF~\cite{yu2020pixelnerf}, which reflect the lower quantitative performance. Both RGBD-Net and pixelNeRF~\cite{yu2020pixelnerf} use extracted features from reference images to render the novel view. However, we train RGBD-Net to produce coarse-to-fine features to predict both color and depth images of the novel views. In case of pixelNeRF~\cite{yu2020pixelnerf}, the single-scale feature maps are not optimized to condition their learned radiance field. Moreover, RGBD-Net generates the whole image significantly faster than those produced by pixelNeRF~\cite{yu2020pixelnerf} due to the fully convolution architecture. Therefore, our method not only produce better novel views but also render them significantly faster than other baselines ~\cite{Nerf,nerfplus,yu2020pixelnerf,trevithick2020grf,wang2021ibrnet}. Comparisons on the average execution time of RGBD-Net and other methods are included in the supplementary material.

Finally, we compare RGBD-Net against recent view synthesis methods~\cite{NPBG,nerfplus} that require per-scene optimization. To compete fairly with these methods, we also fine-tune our pre-trained model on each scene and denote it as  $\text{RGBD-Net}^{\dagger}$ . After finetuning, $\text{RGBD-Net}^{\dagger}$ achieves state-of-the-art results of view synthesis compared to baseline methods which are trained with or without per-scene optimization. In Fig.~\ref{fig_qualNVS}, NeRF++~\cite{nerfplus} fails to render photo-realistic novel views due to some unrealistic noises and blurry edges.
A key component in NeRF~\cite{Nerf} and its variants are the use of the positional encoding~\cite{fourierfeats} which helps generate high frequency details, but positional encoding may also cause unwanted high-frequency artifacts in images, which reduces perceptual quality.

\section{Ablation study} \label{ablation}

\begin{figure}[t]
\begin{center}
  \includegraphics[width=\linewidth]{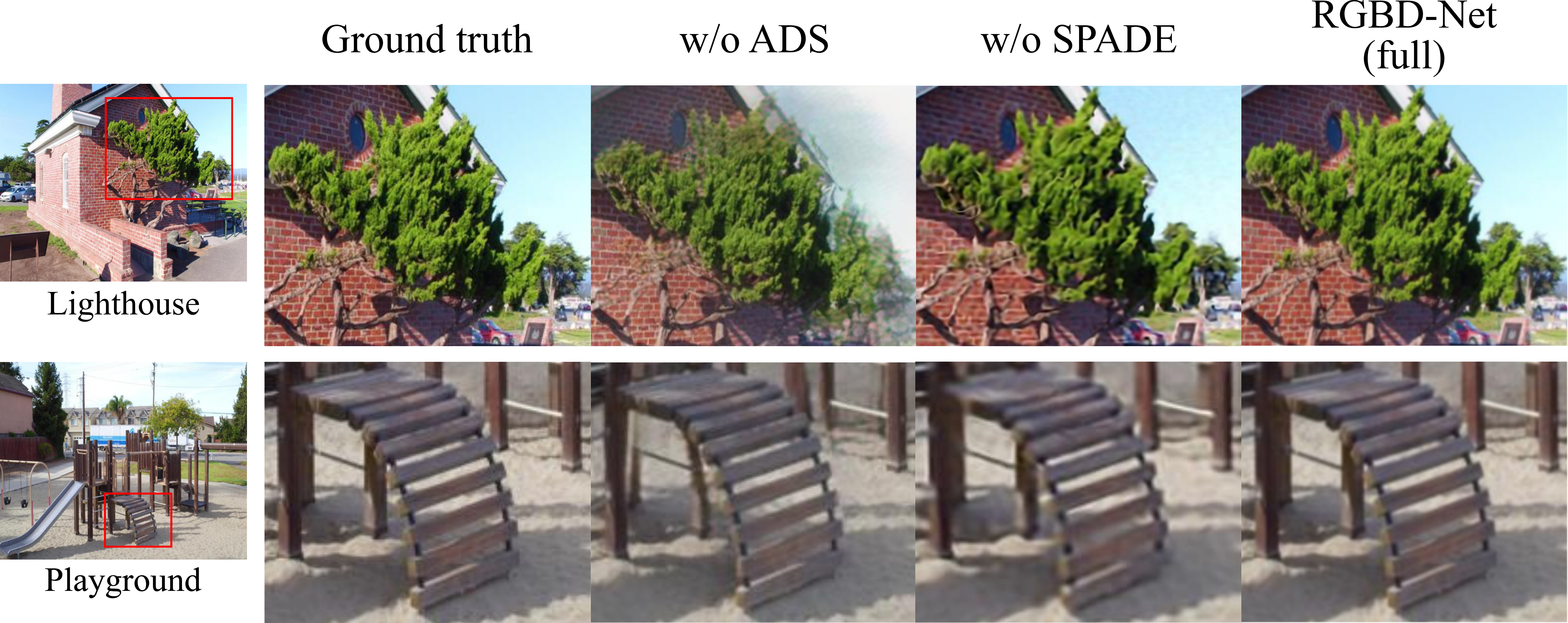}
\end{center}
  \caption{Comparison of the ground-truth with predicted novel views by RGBD-Net without the proposed adaptive depth scaling (ADS), without SPADE Resblocks and the full model.}
\label{fig_arch}
\end{figure}

\begin{table}[t]
\begin{adjustbox}{width=0.95\columnwidth,center}
\centering
\begin{tabular}{@{}lccc@{}}
\toprule
                          & LPIPS$\downarrow$         & SSIM$\uparrow$           & PSNR$\uparrow$           \\ \midrule
No adaptive depth scaling & 0.38          & 0.752          & 18.52          \\
No SPADE Resblocks        & 0.26          & 0.815          & 19.17          \\
No spatial-temporal consistency & 0.21          & 0.879          & 19.94          \\
RGBD-Net (full)           & \textbf{0.16} & \textbf{0.892} & \textbf{21.28} \\ \bottomrule
\end{tabular}
\end{adjustbox}
\caption{
RGBD-Net architecture ablation study.
Reconstruction accuracy of novel view synthesis on  the Tanks and Temples dataset~\cite{TankAndTemples}. 
}
\label{table_arch}
\end{table}

\noindent \textbf{Architecture design.} Table~\ref{table_arch} and Fig.~\ref{fig_arch} summarizes the quantitative and qualitative results on different architecture choices using the test set of the Tanks and Temples dataset \cite{TankAndTemples}. RGBD-Net without the proposed adaptive depth scaling (ADS) does not produce plausible target views as they contain cluttered background and incorrect geometry. Removing this module causes the network to fail to predict the accurate depth map of the target view, leading to visible artifacts and blurriness. Removing SPADE Resblocks also produces degraded novel views compared to the full model of RGBD-Net. The input of each SPADE Resblock is the regressed depth map which contains edge information. As can be seen in the Fig.~\ref{fig_arch}, RGBD-Net without SPADE blocks produces blurry novel views, especially on the edge of objects. Finally, we found that optimizing RGBD-Net to produce a smooth sequence of novel views significantly enhances the overall quality of the independent rendering as can be seen in the Table~\ref{table_arch}. The supplementary video provides a comparison between sequences of novel views with and without the proposed spatial-temporal consistency.

\begin{table}[t]
\begin{adjustbox}{width=0.95\columnwidth,center}
\begin{tabular}{@{}rccccccc@{}}
\toprule
      & \multicolumn{7}{c}{\# of reference images }                               \\ \cmidrule(l){2-8} 
      & 4     & 5     & 6     & 7              & 8     & 9     & 10     \\ \midrule
LPIPS$\downarrow$ & 0.267 & 0.203 & 0.185 & \textbf{0.160} & 0.168 & 0.171 & 0.175 \\
SSIM$\uparrow$  & 0.796 & 0.825 & 0.871 & \textbf{0.892} & 0.890 & 0.890 & 0.887 \\ \bottomrule
\end{tabular}
\end{adjustbox}
\caption{The impact of the number of reference images, measured in terms of novel view reconstruction accuracy on the Tank and Temples dataset \cite{TankAndTemples}.
}
\label{table_inputs}
\end{table}

\noindent \textbf{Number of input views.} In Table~\ref{table_inputs}, we evaluate the performance of our method with an increasing number of source images using the Tanks and Temples \cite{TankAndTemples} dataset. We report both SSIM and LPIPS metrics with the number of source images up to 10. We observe that {RGBD-Net} performs the best with 7 input views and then the results get worse. 

Previous works~\cite{Nerf,LLFF} on view synthesis often render novel views on the densely sampled source images. Therefore, more source views would leads to better performance. In case of large-scaled 3D datasets~\cite{DTU,yao2020blendedmvs,TankAndTemples}, images are captured sparsely around the scene. As can be seen the Fig.~\ref{fig_intro}, some reference views have little overlapping frustum with the target viewpoints. When reference and target poses are far from each other, inaccurate regressed depth maps will lead to less accurate novel views. Therefore, having views close to the target views and having less self-occlusion is essential to synthesize novel views. If it is hard to gather views around the target view, adding more views that have overlapping viewing frustums with the target view is also necessary.

\begin{figure}[t]
\begin{center}
  \includegraphics[width=\linewidth]{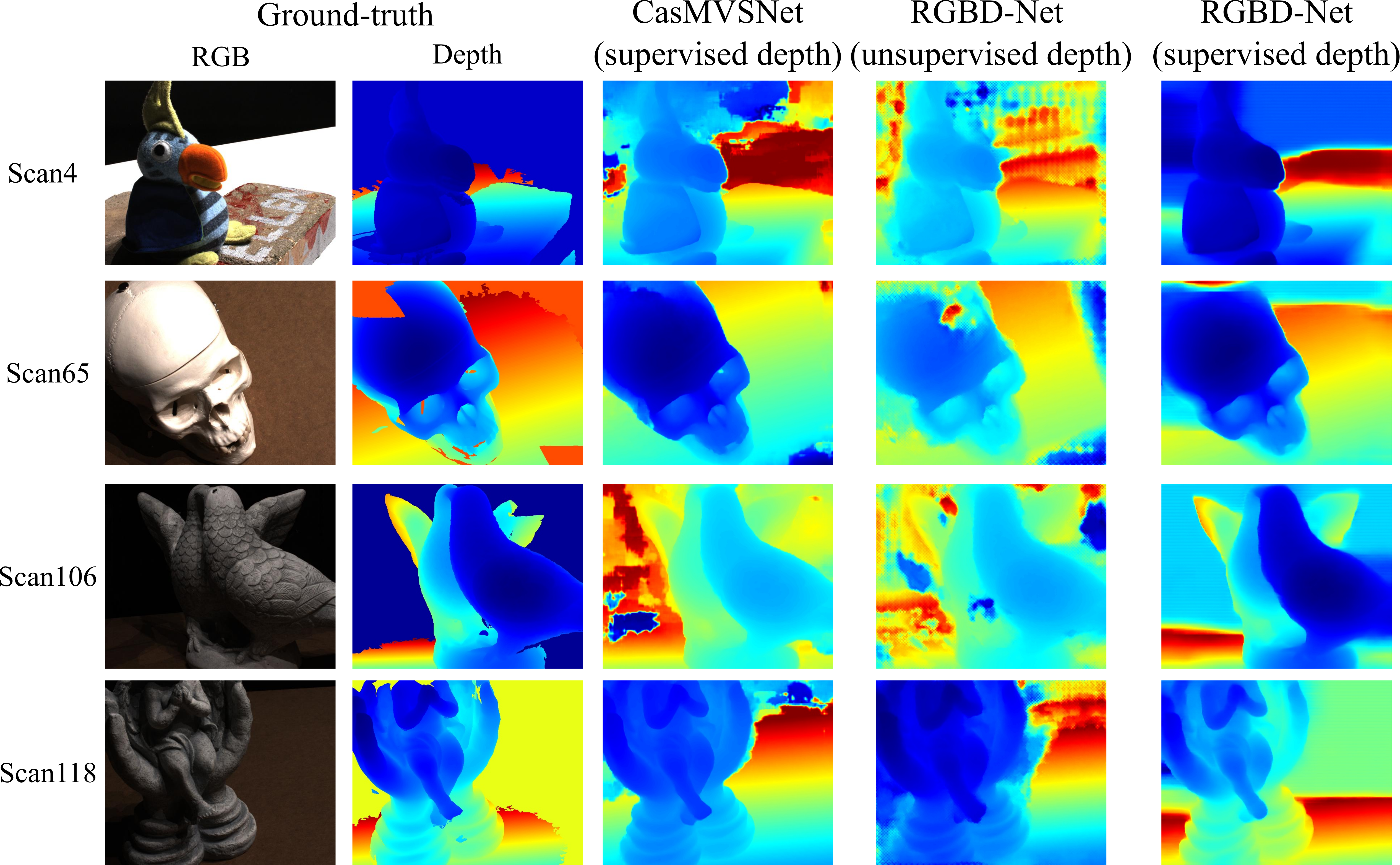}
\end{center}
  \caption{Examples of estimated depth maps using RGBD-Net and CasMVSNet~\cite{CasMVSNet} on the DTU~\cite{DTU} dataset.}
\label{fig_qualDepth}
\end{figure}

\noindent \textbf{Visualizing generated depths.}  In Fig.~\ref{fig_qualDepth}, we show qualitative results on the predicted depth map of the reference camera compared to those produced by the current top-performing MVS method~\cite{CasMVSNet}. Learning-based MVS methods~\cite{MVSNet,CasMVSNet} are trained to predict only depth maps of the given target image and its nearby views. Using the same set of unstructured inputs, RGBD-Net is trying to solve a more challenging problem of predicting both the depth and the color images of the target pose. We observe that our method is able to generate an accurate depth map of the target view without using the reference image as
input. Moreover, we also show that our method performs well on unseen data. In Fig.~\ref{fig_intro} and Fig.~\ref{fig_qualNVS}, our proposed method is able to predict both the depth maps and the color images of the target views and then use them to reconstruct a 3D point cloud on the Tanks and Temples~\cite{TankAndTemples} dataset.

\section{Conclusions}
We presented RGBD-Net, a new method to address the challenging problem of novel view synthesis from a sparse and unstructured set of input images.
Due to its adaptive depth scaling and depth-aware generator network, RGBD-Net is able to produce high-quality depth maps and color images of the target views without per-scene optimization.  
RGBD-Net also achieves unprecedented levels of realism in free-viewpoint video thanks to its novel spatial-temporal module that allows smooth rendering of continuous camera motion.

{\small
\bibliographystyle{ieee_fullname}
\bibliography{egbib}

\begin{thebibliography}{10}\itemsep=-1pt

\bibitem{DTU}
Henrik Aan\ae{}s, Rasmus~Ramsb\o{}l Jensen, George Vogiatzis, Engin Tola, and
  Anders~Bjorholm Dahl.
\newblock Large-scale data for multiple-view stereopsis.
\newblock {\em Int. J. Comput. Vision}, 120(2):153–168, Nov. 2016.

\bibitem{NPBG}
Kara{-}Ali Aliev, Dmitry Ulyanov, and Victor~S. Lempitsky.
\newblock Neural point-based graphics.
\newblock {\em CoRR}, abs/1906.08240, 2019.

\bibitem{intro_warp}
Chris Buehler, Michael Bosse, Leonard McMillan, Steven Gortler, and Michael
  Cohen.
\newblock Unstructured lumigraph rendering.
\newblock In {\em Proceedings of the 28th Annual Conference on Computer
  Graphics and Interactive Techniques}, SIGGRAPH '01, page 425–432, New York,
  NY, USA, 2001. Association for Computing Machinery.

\bibitem{intro_similardepth}
Gaurav Chaurasia, Sylvain Duchene, Olga Sorkine-Hornung, and George Drettakis.
\newblock Depth synthesis and local warps for plausible image-based navigation.
\newblock {\em ACM Trans. Graph.}, 32(3), July 2013.

\bibitem{vggLoss}
Qifeng Chen and Vladlen Koltun.
\newblock Photographic image synthesis with cascaded refinement networks.
\newblock In {\em {IEEE} International Conference on Computer Vision, {ICCV}
  2017, Venice, Italy, October 22-29, 2017}, pages 1520--1529. {IEEE} Computer
  Society, 2017.

\bibitem{pointMSVNet}
Rui Chen, Songfang Han, Jing Xu, and Hao Su.
\newblock Point-based multi-view stereo network.
\newblock In {\em Proceedings of the IEEE International Conference on Computer
  Vision}, pages 1538--1547, 2019.

\bibitem{intro_pixel}
Shenchang~Eric Chen and Lance Williams.
\newblock View interpolation for image synthesis.
\newblock In {\em Proceedings of the 20th Annual Conference on Computer
  Graphics and Interactive Techniques}, SIGGRAPH '93, page 279–288, New York,
  NY, USA, 1993. Association for Computing Machinery.

\bibitem{UCSNet}
Shuo Cheng, Zexiang Xu, Shilin Zhu, Zhuwen Li, Li~Erran Li, Ravi Ramamoorthi,
  and Hao Su.
\newblock Deep stereo using adaptive thin volume representation with
  uncertainty awareness.
\newblock In {\em Proceedings of the IEEE/CVF Conference on Computer Vision and
  Pattern Recognition}, pages 2524--2534, 2020.

\bibitem{EVS}
Inchang Choi, Orazio Gallo, Alejandro Troccoli, Min~H Kim, and Jan Kautz.
\newblock Extreme view synthesis.
\newblock In {\em Proceedings of the IEEE International Conference on Computer
  Vision}, pages 7781--7790, 2019.

\bibitem{PSV}
R.~T. {Collins}.
\newblock A space-sweep approach to true multi-image matching.
\newblock In {\em Proceedings CVPR IEEE Computer Society Conference on Computer
  Vision and Pattern Recognition}, pages 358--363, 1996.

\bibitem{deepview}
John Flynn, Michael Broxton, Paul Debevec, Matthew DuVall, Graham Fyffe, Ryan
  Overbeck, Noah Snavely, and Richard Tucker.
\newblock Deepview: View synthesis with learned gradient descent.
\newblock In {\em Proceedings of the IEEE Conference on Computer Vision and
  Pattern Recognition}, pages 2367--2376, 2019.

\bibitem{fusible}
S. {Galliani}, K. {Lasinger}, and K. {Schindler}.
\newblock Massively parallel multiview stereopsis by surface normal diffusion.
\newblock In {\em 2015 IEEE International Conference on Computer Vision
  (ICCV)}, pages 873--881, 2015.

\bibitem{GAN}
Ian Goodfellow, Jean Pouget-Abadie, Mehdi Mirza, Bing Xu, David Warde-Farley,
  Sherjil Ozair, Aaron Courville, and Yoshua Bengio.
\newblock Generative adversarial nets.
\newblock In Z. Ghahramani, M. Welling, C. Cortes, N.~D. Lawrence, and K.~Q.
  Weinberger, editors, {\em Advances in Neural Information Processing Systems
  27}, pages 2672--2680. Curran Associates, Inc., 2014.

\bibitem{CasMVSNet}
Xiaodong Gu, Zhiwen Fan, Siyu Zhu, Zuozhuo Dai, Feitong Tan, and Ping Tan.
\newblock Cascade cost volume for high-resolution multi-view stereo and stereo
  matching.
\newblock In {\em Proceedings of the IEEE/CVF Conference on Computer Vision and
  Pattern Recognition}, pages 2495--2504, 2020.

\bibitem{depthLosses}
Junjie Hu, Mete Ozay, Yan Zhang, and Takayuki Okatani.
\newblock Revisiting single image depth estimation: Toward higher resolution
  maps with accurate object boundaries.
\newblock In {\em {IEEE} Winter Conference on Applications of Computer Vision,
  {WACV} 2019, Waikoloa Village, HI, USA, January 7-11, 2019}, pages
  1043--1051. {IEEE}, 2019.

\bibitem{DeepMVS}
Po-Han Huang, Kevin Matzen, Johannes Kopf, Narendra Ahuja, and Jia-Bin Huang.
\newblock Deepmvs: Learning multi-view stereopsis.
\newblock In {\em IEEE Conference on Computer Vision and Pattern Recognition
  (CVPR)}, 2018.

\bibitem{STN}
Max Jaderberg, Karen Simonyan, Andrew Zisserman, and Koray Kavukcuoglu.
\newblock Spatial transformer networks.
\newblock In {\em Proceedings of the 28th International Conference on Neural
  Information Processing Systems - Volume 2}, NIPS'15, page 2017–2025,
  Cambridge, MA, USA, 2015. MIT Press.

\bibitem{Kalantari}
Nima~Khademi Kalantari, Ting-Chun Wang, and Ravi Ramamoorthi.
\newblock Learning-based view synthesis for light field cameras.
\newblock {\em ACM Trans. Graph.}, 35(6), Nov. 2016.

\bibitem{MSGGAN}
Animesh Karnewar and Oliver Wang.
\newblock Msg-gan: Multi-scale gradients for generative adversarial networks.
\newblock In {\em Proceedings of the IEEE/CVF Conference on Computer Vision and
  Pattern Recognition}, pages 7799--7808, 2020.

\bibitem{styleGAN2}
Tero Karras, Samuli Laine, Miika Aittala, Janne Hellsten, Jaakko Lehtinen, and
  Timo Aila.
\newblock Analyzing and improving the image quality of {StyleGAN}.
\newblock In {\em Proc. CVPR}, 2020.

\bibitem{TankAndTemples}
Arno Knapitsch, Jaesik Park, Qian-Yi Zhou, and Vladlen Koltun.
\newblock Tanks and temples: Benchmarking large-scale scene reconstruction.
\newblock {\em ACM Transactions on Graphics}, 36(4), 2017.

\bibitem{Pulsar}
Christoph Lassner and Michael Zollhöfer.
\newblock Pulsar: Efficient sphere-based neural rendering, 2020.

\bibitem{intro_ray}
Marc Levoy and Pat Hanrahan.
\newblock Light field rendering.
\newblock In {\em Proceedings of the 23rd Annual Conference on Computer
  Graphics and Interactive Techniques}, SIGGRAPH '96, page 31–42, New York,
  NY, USA, 1996. Association for Computing Machinery.

\bibitem{crowdsampling}
Zhengqi Li, Wenqi Xian, Abe Davis, and Noah Snavely.
\newblock Crowdsampling the plenoptic function.
\newblock In {\em Proc. European Conference on Computer Vision (ECCV)}, 2020.

\bibitem{FPN}
Tsung-Yi Lin, Piotr Doll{\'a}r, Ross Girshick, Kaiming He, Bharath Hariharan,
  and Serge Belongie.
\newblock Feature pyramid networks for object detection.
\newblock In {\em Proceedings of the IEEE conference on computer vision and
  pattern recognition}, pages 2117--2125, 2017.

\bibitem{SparseVoxelFields}
Lingjie Liu, Jiatao Gu, Kyaw~Zaw Lin, Tat-Seng Chua, and Christian Theobalt.
\newblock Neural sparse voxel fields.
\newblock {\em NeurIPS}, 2020.

\bibitem{neuralVolumes}
Stephen Lombardi, Tomas Simon, Jason Saragih, Gabriel Schwartz, Andreas
  Lehrmann, and Yaser Sheikh.
\newblock Neural volumes: Learning dynamic renderable volumes from images.
\newblock {\em ACM Trans. Graph.}, 38(4), July 2019.

\bibitem{LLFF}
Ben Mildenhall, Pratul~P. Srinivasan, Rodrigo Ortiz-Cayon, Nima~Khademi
  Kalantari, Ravi Ramamoorthi, Ren Ng, and Abhishek Kar.
\newblock Local light field fusion: Practical view synthesis with prescriptive
  sampling guidelines.
\newblock {\em ACM Transactions on Graphics (TOG)}, 2019.

\bibitem{Nerf}
Ben Mildenhall, Pratul~P. Srinivasan, Matthew Tancik, Jonathan~T. Barron, Ravi
  Ramamoorthi, and Ren Ng.
\newblock Nerf: Representing scenes as neural radiance fields for view
  synthesis.
\newblock In {\em ECCV}, 2020.

\bibitem{xfield}
Hans-Peter~Seidel Mojtaba~Bemana, Karol~Myszkowski and Tobias Ritschel.
\newblock X-fields: Implicit neural view-, light- and time-image interpolation.
\newblock {\em ACM Transactions on Graphics (Proc. SIGGRAPH Asia 2020)}, 39(6),
  2020.

\bibitem{neuralrep1}
Michael Niemeyer, Lars Mescheder, Michael Oechsle, and Andreas Geiger.
\newblock Differentiable volumetric rendering: Learning implicit 3d
  representations without 3d supervision.
\newblock In {\em Proceedings IEEE Conf. on Computer Vision and Pattern
  Recognition (CVPR)}, 2020.

\bibitem{spade}
Taesung Park, Ming-Yu Liu, Ting-Chun Wang, and Jun-Yan Zhu.
\newblock Semantic image synthesis with spatially-adaptive normalization.
\newblock In {\em Proceedings of the IEEE Conference on Computer Vision and
  Pattern Recognition}, 2019.

\bibitem{Pytorch}
Adam Paszke, Sam Gross, Francisco Massa, Adam Lerer, James Bradbury, Gregory
  Chanan, Trevor Killeen, Zeming Lin, Natalia Gimelshein, Luca Antiga, Alban
  Desmaison, Andreas Kopf, Edward Yang, Zachary DeVito, Martin Raison, Alykhan
  Tejani, Sasank Chilamkurthy, Benoit Steiner, Lu Fang, Junjie Bai, and Soumith
  Chintala.
\newblock Pytorch: An imperative style, high-performance deep learning library.
\newblock In {\em Advances in Neural Information Processing Systems 32}, pages
  8026--8037. Curran Associates, Inc., 2019.

\bibitem{intro_soft3d}
Eric Penner and Li Zhang.
\newblock Soft 3d reconstruction for view synthesis.
\newblock {\em ACM Trans. Graph.}, 36(6), Nov. 2017.

\bibitem{alpha}
Thomas Porter and Tom Duff.
\newblock Compositing digital images.
\newblock In {\em Proceedings of the 11th Annual Conference on Computer
  Graphics and Interactive Techniques}, SIGGRAPH '84, page 253–259, New York,
  NY, USA, 1984. Association for Computing Machinery.

\bibitem{FVS}
Gernot Riegler and Vladlen Koltun.
\newblock Free view synthesis.
\newblock In {\em European Conference on Computer Vision}, 2020.

\bibitem{stable}
Gernot Riegler and Vladlen Koltun.
\newblock Stable view synthesis.
\newblock {\em arXiv preprint arXiv:2011.07233}, 2020.

\bibitem{colmap}
Johannes~Lutz Sch\"{o}nberger and Jan-Michael Frahm.
\newblock Structure-from-motion revisited.
\newblock In {\em Conference on Computer Vision and Pattern Recognition
  (CVPR)}, 2016.

\bibitem{ConvLSTM}
Xingjian Shi, Zhourong Chen, Hao Wang, Dit-Yan Yeung, Wai-kin Wong, and
  Wang-chun Woo.
\newblock Convolutional lstm network: A machine learning approach for
  precipitation nowcasting.
\newblock In {\em Proceedings of the 28th International Conference on Neural
  Information Processing Systems - Volume 1}, NIPS'15, page 802–810,
  Cambridge, MA, USA, 2015. MIT Press.

\bibitem{VR}
N.~K. {Shukia}, S. {Sengupta}, and M. {Chakraborty}.
\newblock Intermediate view synthesis in wide-baseline stereoscopic video for
  immersive telepresence.
\newblock In {\em The 2nd European Workshop on the Integration of Knowledge,
  Semantics and Digital Media Technology, 2005. EWIMT 2005. (Ref. No.
  2005/11099)}, pages 83--88, 2005.

\bibitem{Deepvoxels}
Vincent Sitzmann, Justus Thies, Felix Heide, Matthias Nie{\ss}ner, Gordon
  Wetzstein, and Michael Zollh{\"o}fer.
\newblock Deepvoxels: Learning persistent 3d feature embeddings.
\newblock In {\em Proc. Computer Vision and Pattern Recognition (CVPR), IEEE},
  2019.

\bibitem{neuralrep3}
Vincent Sitzmann, Michael Zollh{\"o}fer, and Gordon Wetzstein.
\newblock Scene representation networks: Continuous 3d-structure-aware neural
  scene representations.
\newblock In {\em Advances in Neural Information Processing Systems}, pages
  1121--1132, 2019.

\bibitem{MPI2}
P.~P. {Srinivasan}, R. {Tucker}, J.~T. {Barron}, R. {Ramamoorthi}, R. {Ng}, and
  N. {Snavely}.
\newblock Pushing the boundaries of view extrapolation with multiplane images.
\newblock In {\em 2019 IEEE/CVF Conference on Computer Vision and Pattern
  Recognition (CVPR)}, pages 175--184, 2019.

\bibitem{fourierfeats}
Matthew Tancik, Pratul~P. Srinivasan, Ben Mildenhall, Sara Fridovich-Keil,
  Nithin Raghavan, Utkarsh Singhal, Ravi Ramamoorthi, Jonathan~T. Barron, and
  Ren Ng.
\newblock Fourier features let networks learn high frequency functions in low
  dimensional domains.
\newblock {\em NeurIPS}, 2020.

\bibitem{report}
A. Tewari, O. Fried, J. Thies, V. Sitzmann, S. Lombardi, K. Sunkavalli, R.
  Martin-Brualla, T. Simon, J. Saragih, M. Nie{\ss}ner, R. Pandey, S. Fanello,
  G. Wetzstein, J.-Y. Zhu, C. Theobalt, M. Agrawala, E. Shechtman, D.~B
  Goldman, and M. Zollh{\"o}fer.
\newblock {State of the Art on Neural Rendering}.
\newblock {\em Computer Graphics Forum (EG STAR 2020)}, 2020.

\bibitem{neuralrep4}
Justus Thies, Michael Zollh{\"o}fer, and Matthias Nie{\ss}ner.
\newblock Deferred neural rendering: Image synthesis using neural textures.
\newblock {\em ACM Transactions on Graphics (TOG)}, 38(4):1--12, 2019.

\bibitem{trevithick2020grf}
Alex Trevithick and Bo Yang.
\newblock Grf: Learning a general radiance field for 3d scene representation
  and rendering.
\newblock {\em arXiv preprint arXiv:2010.04595}, 2020.

\bibitem{MPI3}
Richard Tucker and Noah Snavely.
\newblock Single-view view synthesis with multiplane images.
\newblock In {\em The IEEE Conference on Computer Vision and Pattern
  Recognition (CVPR)}, June 2020.

\bibitem{wang2021ibrnet}
Qianqian Wang, Zhicheng Wang, Kyle Genova, Pratul Srinivasan, Howard Zhou,
  Jonathan~T. Barron, Ricardo Martin-Brualla, Noah Snavely, and Thomas
  Funkhouser.
\newblock Ibrnet: Learning multi-view image-based rendering.
\newblock {\em arXiv preprint arXiv:2102.13090}, 2021.

\bibitem{hybridnerf}
Ziyan Wang, Timur Bagautdinov, Stephen Lombardi, Tomas Simon, Jason Saragih,
  Jessica Hodgins, and Michael Zollhöfer.
\newblock Learning compositional radiance fields of dynamic human heads, 2020.

\bibitem{synsin}
Olivia Wiles, Georgia Gkioxari, Richard Szeliski, and Justin Johnson.
\newblock Synsin: End-to-end view synthesis from a single image.
\newblock In {\em Proceedings of the IEEE/CVF Conference on Computer Vision and
  Pattern Recognition}, pages 7467--7477, 2020.

\bibitem{MVSNet}
Yao Yao, Zixin Luo, Shiwei Li, Tian Fang, and Long Quan.
\newblock Mvsnet: Depth inference for unstructured multi-view stereo.
\newblock {\em European Conference on Computer Vision (ECCV)}, 2018.

\bibitem{yao2020blendedmvs}
Yao Yao, Zixin Luo, Shiwei Li, Jingyang Zhang, Yufan Ren, Lei Zhou, Tian Fang,
  and Long Quan.
\newblock Blendedmvs: A large-scale dataset for generalized multi-view stereo
  networks.
\newblock {\em Computer Vision and Pattern Recognition (CVPR)}, 2020.

\bibitem{PVAMVSNet}
Hongwei Yi, Zizhuang Wei, Mingyu Ding, Runze Zhang, Yisong Chen, Guoping Wang,
  and Yu-Wing Tai.
\newblock Pyramid multi-view stereo net with self-adaptive view aggregation.
\newblock {\em arXiv preprint arXiv:1912.03001}, 2019.

\bibitem{yu2020pixelnerf}
Alex Yu, Vickie Ye, Matthew Tancik, and Angjoo Kanazawa.
\newblock pixelnerf: Neural radiance fields from one or few images, 2020.

\bibitem{nerfplus}
Kai Zhang, Gernot Riegler, Noah Snavely, and Vladlen Koltun.
\newblock Nerf++: Analyzing and improving neural radiance fields.
\newblock {\em arXiv:2010.07492}, 2020.

\bibitem{LPIPS}
Richard Zhang, Phillip Isola, Alexei~A Efros, Eli Shechtman, and Oliver Wang.
\newblock The unreasonable effectiveness of deep features as a perceptual
  metric.
\newblock In {\em CVPR}, 2018.

\bibitem{StereoMag}
Tinghui Zhou, Richard Tucker, John Flynn, Graham Fyffe, and Noah Snavely.
\newblock Stereo magnification: Learning view synthesis using multiplane
  images.
\newblock In {\em SIGGRAPH}, 2018.

\bibitem{sean}
Peihao Zhu, Rameen Abdal, Yipeng Qin, and Peter Wonka.
\newblock Sean: Image synthesis with semantic region-adaptive normalization.
\newblock In {\em IEEE/CVF Conference on Computer Vision and Pattern
  Recognition (CVPR)}, June 2020.

\end{thebibliography}
}

\clearpage

\begin{center}
    \textbf{\Large Appendix of RGBD-Net}    
\end{center}

\setcounter{section}{0}
\renewcommand\thesection{\Alph{section}}
\renewcommand\thesubsection{\thesection.\arabic{subsection}}

\begin{table*}[ht]
\centering
\resizebox{\textwidth}{!}{%
\begin{tabular}{lcccccccccccc}
\hline
\multirow{2}{*}{} & \multicolumn{3}{c}{DTU \cite{DTU}} & \multicolumn{9}{c}{Tanks and Temples \cite{TankAndTemples}}                                 \\ \cline{2-13} 
 &
  \begin{tabular}[c]{@{}c@{}}Acc.$\downarrow$ \\ (mm)\end{tabular} &
  \begin{tabular}[c]{@{}c@{}}Comp.$\downarrow$ \\ (mm)\end{tabular} &
  \begin{tabular}[c]{@{}c@{}}Overall$\downarrow$ \\ (mm)\end{tabular} &
  Mean$\uparrow$ &
  Family$\uparrow$ &
  Francis$\uparrow$ &
  Horse$\uparrow$ &
  Lighthouse$\uparrow$ &
  M60$\uparrow$ &
  Panther$\uparrow$ &
  Playground$\uparrow$ &
  Train$\uparrow$ \\ \hline
MVSNet \cite{MVSNet}            & 0.456  & 0.646  & 0.551 & 43.48 & 55.99 & 28.55 & 25.07 & 50.79 & 53.96 & 50.86 & 47.90 & 34.69 \\
Point-MVSNet \cite{pointMSVNet}      & 0.361  & 0.421  & 0.391 & 48.27 & 61.79 & 41.15 & 34.20 & 50.79 & 51.97 & 50.85 & 52.38 & 43.06 \\
PVA-MVSNet \cite{PVAMVSNet}        & 0.352  & 0.414  & 0.383 & 54.46 & 69.36 & 46.80 & 46.01 & 55.74 & 57.23 & 54.75 & 56.70 & 49.06 \\
UCSNet \cite{UCSNet}            & 0.330  & 0.392  & 0.361 & 54.83 & 76.09 & 53.16 & 43.04 & 54.00 & 55.60 & 51.49 & 57.38 & 47.89 \\
CasMVSNet \cite{CasMVSNet}        & 0.325  & 0.385  & 0.355 & 56.84 & 76.37 & 58.45 & 46.26 & 55.81 & 56.11 & 54.06 & 58.18 & 49.51 \\
{$\text{RGBD-Net}^*$}            & 0.334  & 0.390  & 0.362 & 55.57 & 76.82 & 53.84 & 44.29 & 55.02 & 55.98 & 52.78 & 58.63 & 47.25 \\
\textbf{RGBD-Net} &
  \textbf{0.320} &
  \textbf{0.381} &
  \textbf{0.349} &
  \textbf{59.32} &
  \textbf{77.01} &
  \textbf{60.25} &
  \textbf{47.09} &
  \textbf{63.45} &
  \textbf{62.19} &
  \textbf{55.16} &
  \textbf{59.27} &
  \textbf{50.19} \\ \hline
\end{tabular}%
}
\caption{Point cloud accuracy on the DTU test \cite{DTU} and 
Tanks and Temples \cite{TankAndTemples}
\textit{intermediate}  datasets.}
\label{table_quant_depth}
\end{table*}

\begin{figure}[ht]
\begin{center}
  \includegraphics[width=\linewidth]{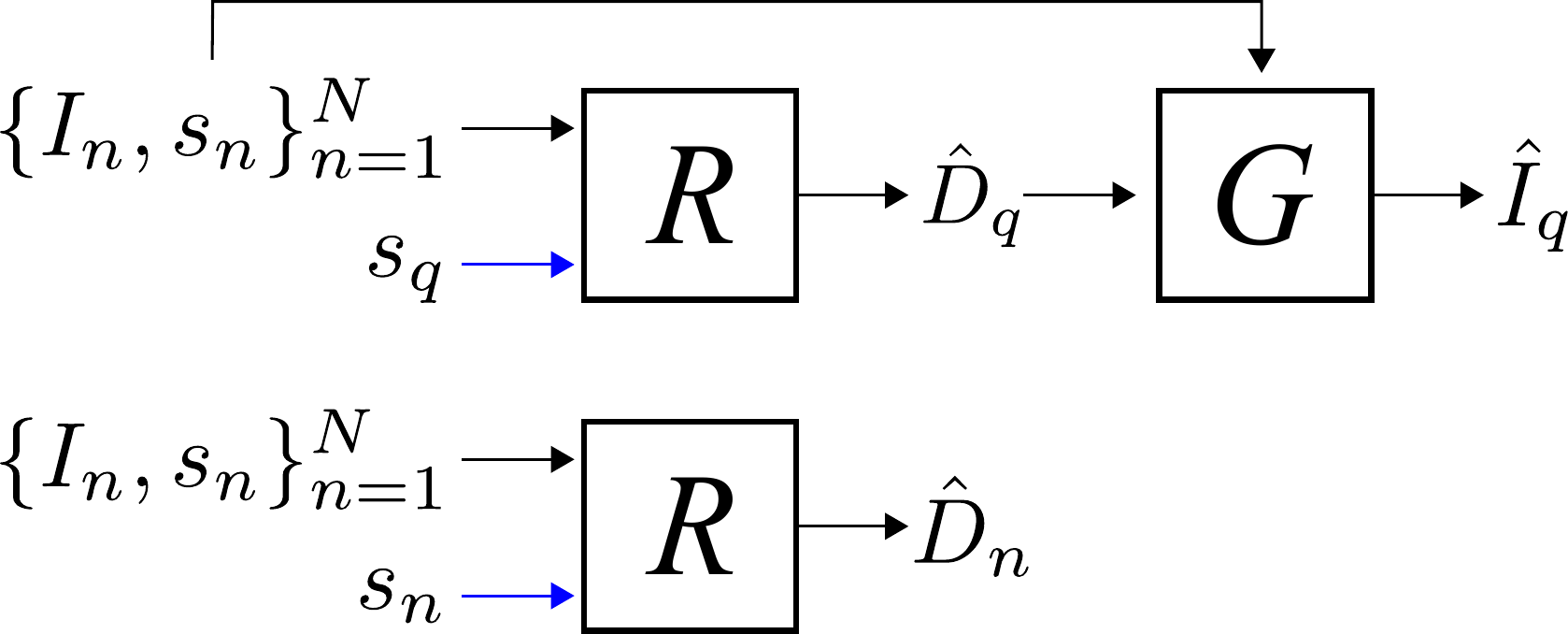}
\end{center}
  \caption{The illustration of how RGBD-Net produces the RGB image $\hat{I}_q$, depth map $\hat{D}_q$ at the target view $s_q$ and the depth map $\hat{D}_n$ of the reference pose $s_n$ in the testing time. Blue arrow indicates the input pose of the depth regression network $R$.}
  \label{fig_point}
\end{figure}

\section{Multi-view stereo evaluation}\label{depth_evaluation}
\begin{figure*}[t]
\begin{center}
  \includegraphics[width=\linewidth]{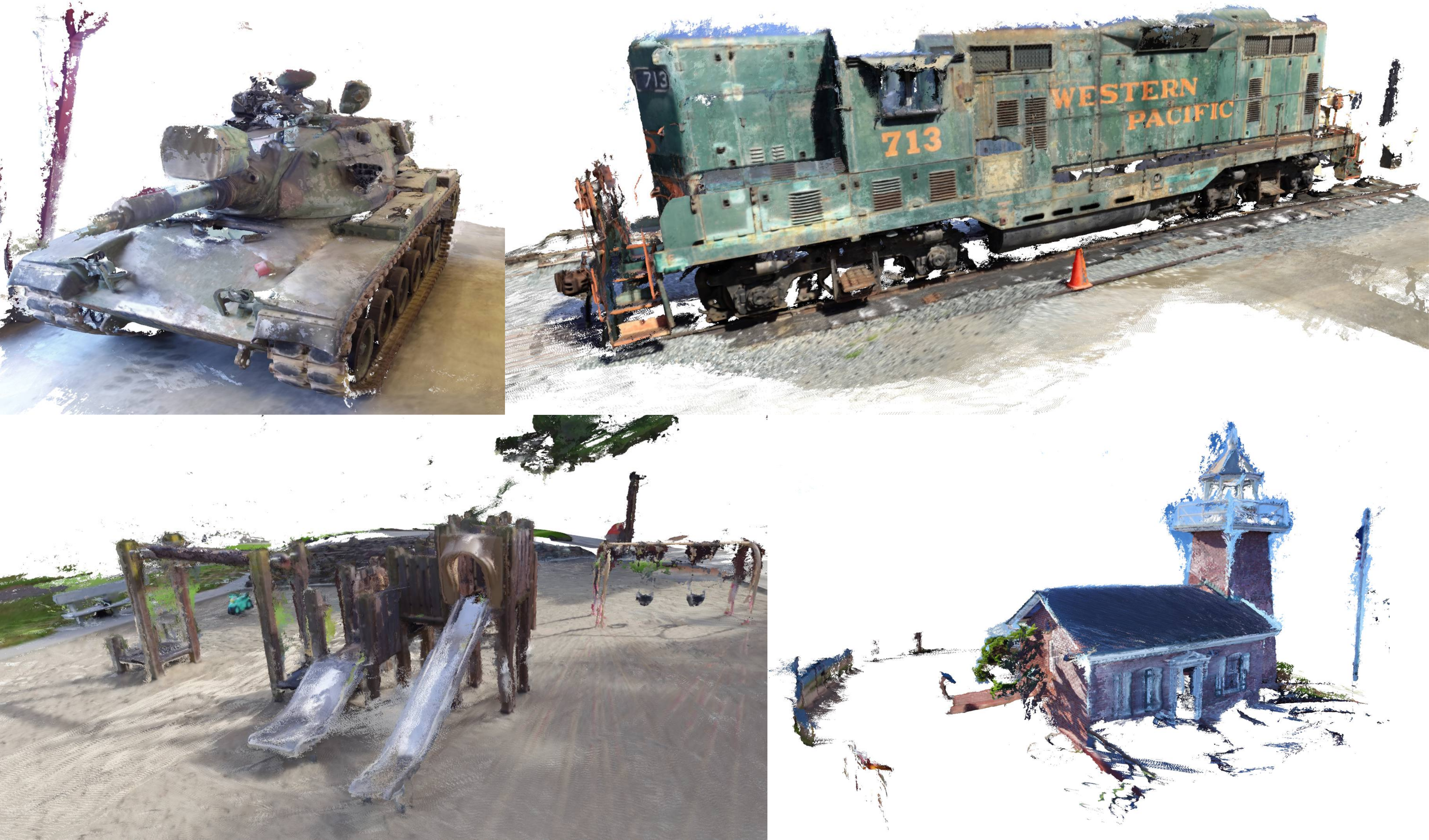}
\end{center}
  \caption{Pointcloud results of RGBD-Net  on the \textit{intermediate} set of Tanks and Temples \cite{yao2020blendedmvs} dataset.}
\label{fig_pcTT}
\end{figure*}

\begin{figure*}[t]
\begin{center}
  \includegraphics[width=\linewidth]{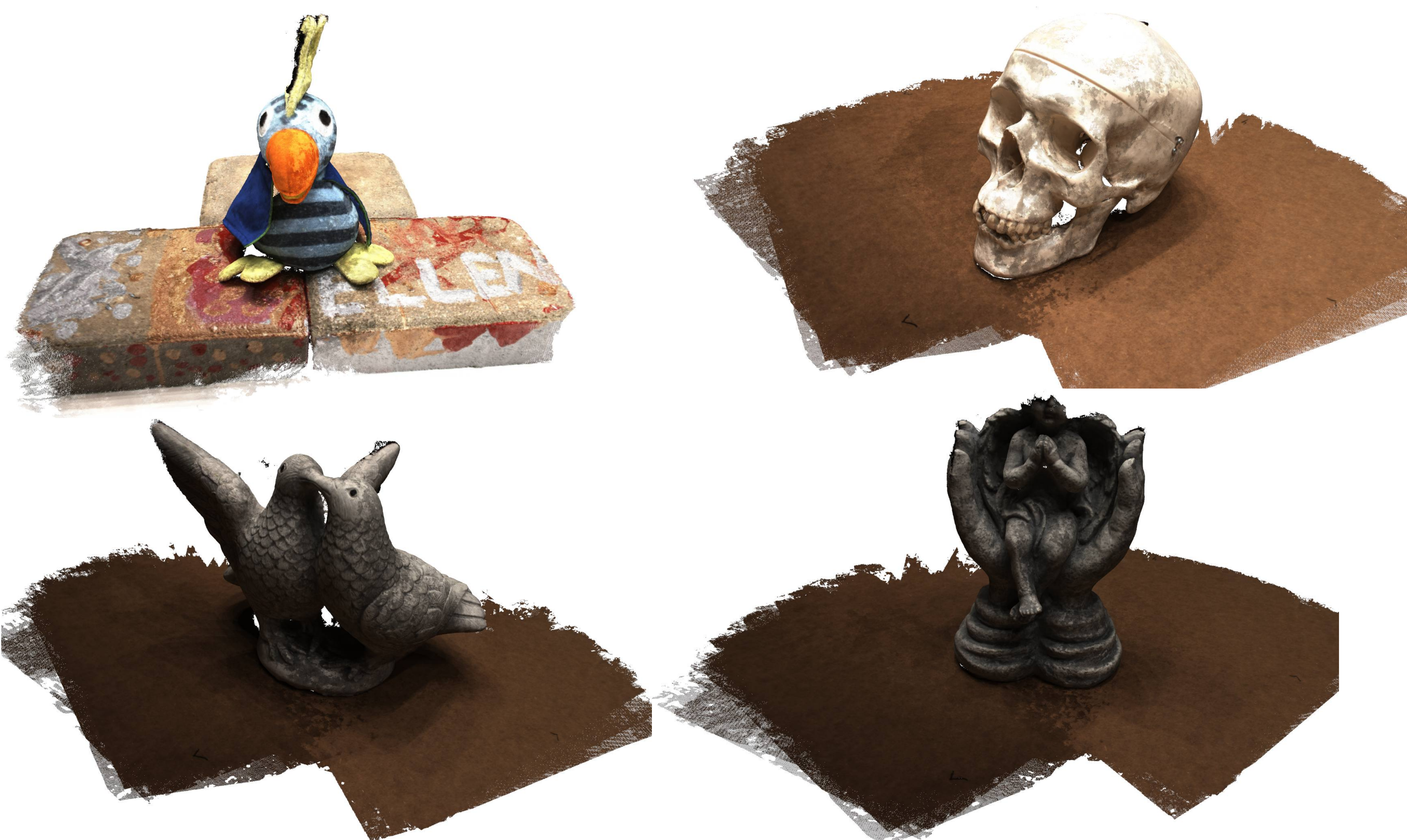}
\end{center}
  \caption{Pointcloud results of RGBD-Net  on the DTU test set \cite{DTU}.}
\label{fig_pcDTU}
\end{figure*}

\begin{figure*}[t]
\begin{center}
  \includegraphics[width=\linewidth]{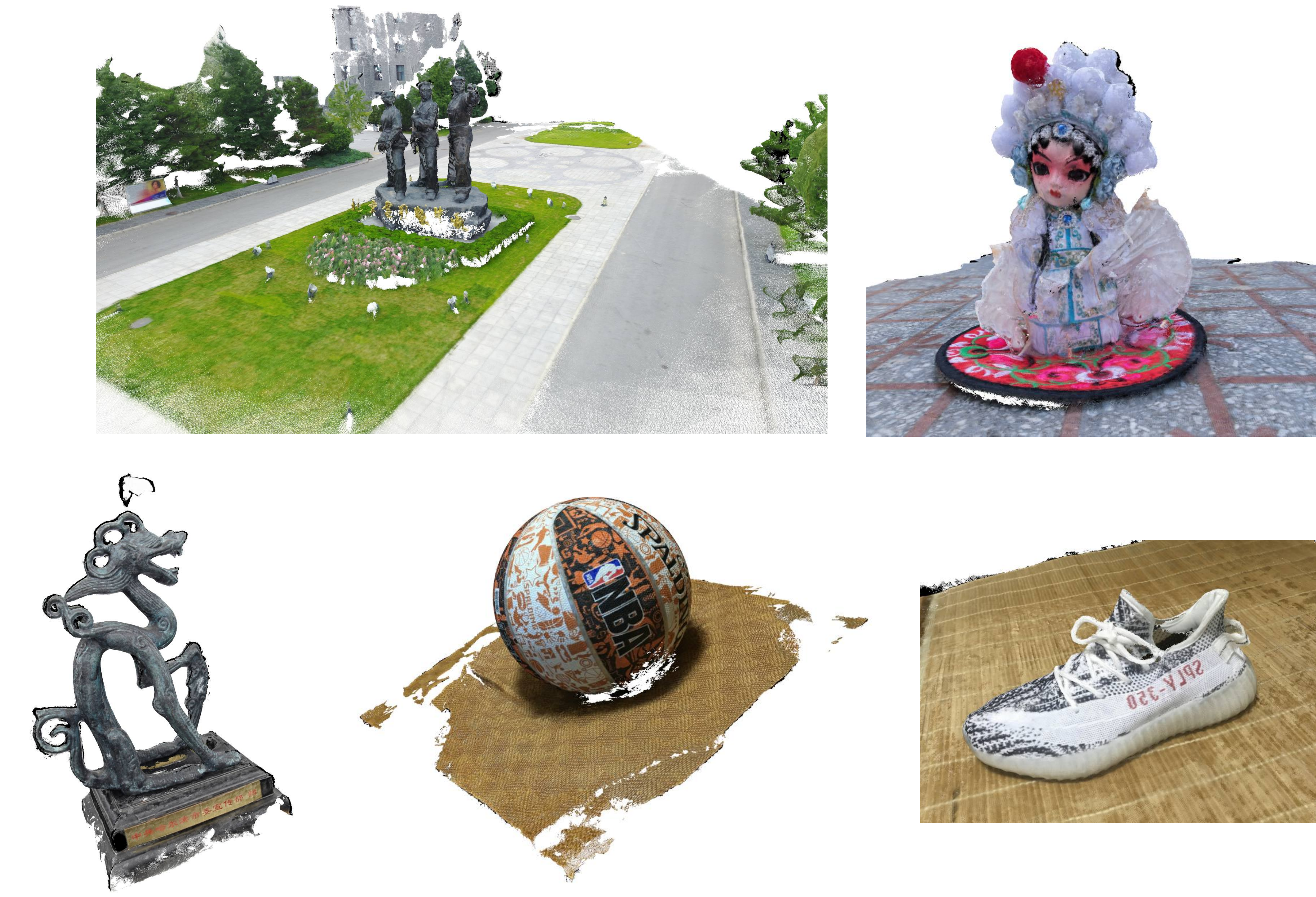}
\end{center}
  \caption{Pointcloud results of RGBD-Net  on the BlendedMVS test set \cite{yao2020blendedmvs}.}
\label{fig_pcBlend}
\end{figure*}

\noindent\textbf{Pointcloud generation.} Similar to previous works~\cite{MVSNet,CasMVSNet} on MVS, we apply depth map filter and fusion approach~\cite{fusible} to merge all predicted novel views and depth maps into a unified pointcloud output. In the testing time, we predict the depth map $\hat{D}_n$ of each reference view $I_n$ using the trained depth regression network $R$ as can be seen in Fig.~\ref{fig_point}. We use the union set $\Psi=\{\hat{I}_q,\hat{D}_q\}_{q=1}^Q \cup \{I_n,\hat{D}_n\}_{n=1}^N$ of the estimated novel views and depth maps as inputs for pointcloud generation\footnote{We use the predictions of the final scale so superscript $k$ is omitted.}. We apply two-step depth map filtering strategy to remove outliers. Finally, median depth map fusion is applied to refine all depth maps. The 3D pointcloud is obtained by projecting all refined depth maps into the 3D space.
\\
\noindent\textbf{Baselines.} 
We evaluate the predicted depth maps produced by RGBD-Net against MVS methods \cite{MVSNet,pointMSVNet,PVAMVSNet,UCSNet,CasMVSNet} on the DTU \cite{DTU} test set and the \textit{intermediate} set of Tanks and Temples \cite{TankAndTemples} dataset. We use fusible \cite{fusible} as the post-processing step to reconstruct the 3D point cloud of the scene. Therefore, a more accurate and consistently estimated depth map would lead to better performance in 3D reconstruction. We note that  our problem is a more challenging case since RGBD-Net predicts both the target depth  maps  and  color  images  using  only  the  reference views. Whereas, all other MVS methods predict only the depth maps at the reference poses while using the reference images as input.
\\
\noindent\textbf{Metrics.} We calculate the mean \textit{accuracy}, \textit{completeness} and \textit{overall} using the evaluation code provided by \cite{DTU}. The average of mean accuracy and completeness represent the reconstruction quality. Moreover, we also calculate the mean F-score on the Tanks and Temples \cite{TankAndTemples} dataset.  
\\
\noindent\textbf{Results.} As can be seen in Table \ref{table_quant_depth}, the proposed method trained without the depth loss ({$\text{RGBD-Net}^*$}) shows competitive performance with other MVS baselines. Using the ground-truth depth loss, our full model ({RGBD-Net}) achieves the best mean F-score on Tanks and Temples and the best overall distance on DTU. We observe that our method is able to generate accurate depth map of the target view without using the reference image as input. Moreover, we also show that our method performs well on unseen data. Fig. \ref{fig_pcTT}, \ref{fig_pcDTU} and \ref{fig_pcBlend} show more examples of generated pointclouds using the proposed RGBD-Net on the Tanks and Temples \cite{TankAndTemples}, DTU \cite{DTU} and BlendedMVS \cite{yao2020blendedmvs} datasets, respectively. 

\begin{figure*}[ht]
\begin{center}
  \includegraphics[width=\linewidth]{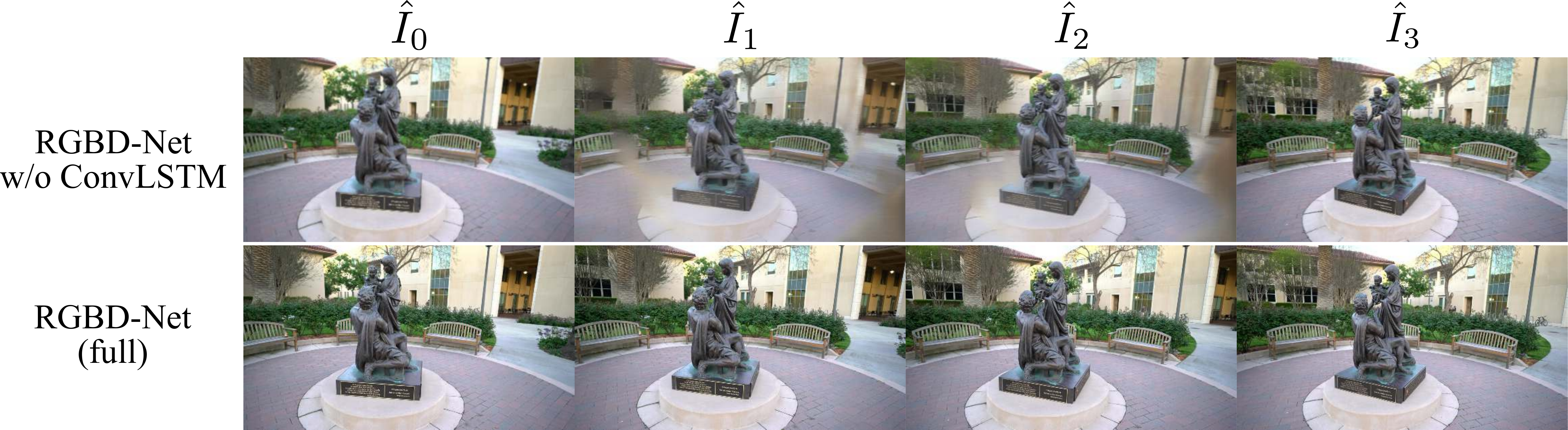}
\end{center}
  \caption{A generated sequence $\{\hat{I}_q\}$ of novel views produced by RGBD-Net with and without the spatial-temporal consistency module.}
  \label{fig_temporal}
\end{figure*}

\section{Spatial-temporal consistency} \label{temporal}
We found that optimizing RGBD-Net to produce a smooth sequence of novel views significantly enhances the overall quality of the independent rendering as
can be seen in the Table 2 of the main paper. Thus, we provide qualitative results of a sequence of novel views produces by RGBD-Net with and without the proposed spatial-temporal consistency module.
As can be seen in Fig.~\ref{fig_temporal}, the predicted novel views include significant artifacts near the boundary and also not very temporally consistent due to independent renderings at each novel viewpoint using 2D/3D U-Net. The proposed depth-aware ConvLSTM allows RGBD-Net to retain the hidden state from previous steps and refine the current generated novel view. 

\section{Execution time} \label{time}

\begin{table}[ht]
\centering
\begin{adjustbox}{width=\columnwidth,center}
\begin{tabular}{ccccccc}
\hline
Methods                                                       & pixelNeRF \cite{yu2020pixelnerf} & LLFF \cite{LLFF} & FVS \cite{FVS} & NPBG \cite{NPBG} & NeRF++ \cite{nerfplus} & \begin{tabular}[c]{@{}c@{}}RGBD-Net\\ (ours)\end{tabular} \\ \hline
\begin{tabular}[c]{@{}c@{}}Avg time\\ (ms / img)\end{tabular} & 9262 & 651  & 541 & 352  & 8153 & \textbf{156}                                                       \\ \hline
\end{tabular}
\end{adjustbox}
\caption{Comparisons on the average execution time of RGBD-Net and other view synthesis methods.}
\label{table_time}
\end{table}

In Table.~\ref{table_time}, we report the average execution times to synthesize a novel image between RGBD-Net with different methods. In all experiments, we synthesize the novel image with the size of 640 $\times$ 512 pixels using 4 reference views on the T$\&$T datasets. Notice that, RGBD-Net is 52 times faster than the current state-of-the-art neural rendering method NeRF++ \cite{nerfplus}. Moreover, our method also run faster than other image-based rendering techniques while maintaining superior performance.

\section{Implementation details} \label{implementation}
\noindent\textbf{Training.} We trained RGBD-Net on the DTU dataset for 25 epochs and subsequently finetuned it on the BlendedMVS dataset for 30 epochs. The models were trained with the Adam optimizer with a batch size of 4. In both training sets, the input and output reference views have the same image size of 640 $\times$ 512 pixels and we set the number of reference views to $N = 4$. To balance between accuracy and efficiency, we adopt a three-scale ($K=3$) depth regression network $R$. Accordingly, the spatial resolution of extracted feature maps $F^k_n$ is set to 1/16, 1/4 and 1 of the original image size.
\\
\noindent\textbf{Adaptive depth scaling.} In Section 3.1 of the paper, our proposed adaptive depth scalin depends on the initial number of hypothesis depth plane $M_1$, minimum depth value $d^1_{min}$ and depth interval $\Delta_1$. In all experiments, we use the same the number of hypothesis depth plane $M_1 = 48$ at the first stage.

The DTU~\cite{DTU} dataset is captured in a controlled environment so we follow previous works~\cite{MVSNet,CasMVSNet} on MVS and set $d^1_{min} = 425$ and $\Delta_1 = 10.6$. The BlendedMVS~\cite{yao2020blendedmvs} and Tanks and Temples (T$\&$T) \cite{TankAndTemples} contain large-scaled scenes which have variety of depth ranges. Some depth ranges are from 0.1 to 2 or from 10 to 100. Note that these numbers are not the absolute distances in some known units. Therefore, we scale those depth ranges roughly to the same scale from 100 to 1000. In that way, $d^1_{min}$ and $\Delta_1$ are chosen differently based on the per-scene scaling factor $f$. Moreover, the scaled depth ranges in BlendedMVS and T$\&$T datasets become approximately similar to the depth range of the DTU dataset. This allows us to train RGBD-Net on both DTU and BlendedMVS datasets and test the generalization of the model on the T$\&$T dataset.
\\
\noindent\textbf{Depth map regression.} Inspired by current learning-based MVS methods~\cite{MVSNet,CasMVSNet}, the predicted depth map is regressed from the probability volume via the \textit{soft-argmax} operation. We denote the probability volume over all the $M_k$ depth hypothesis as $V^k$. The predicted depth value $\hat{D}^k_q(p)$ of each pixel $p$ is defined as follows:
\begin{equation}
    \hat{D}^k_q(p)=\sum_{i=1}^{M_k} d^k_i V^k_i(p)
\end{equation}
where $V^k_i(p)$ is the predicted probability of the depth plane $d^k_i$ for the pixel $p$ at the $k$-th scale.

\section{Additional qualitative results} \label{Qual_results}
\begin{figure*}[t]
\begin{center}
  \includegraphics[width=\linewidth]{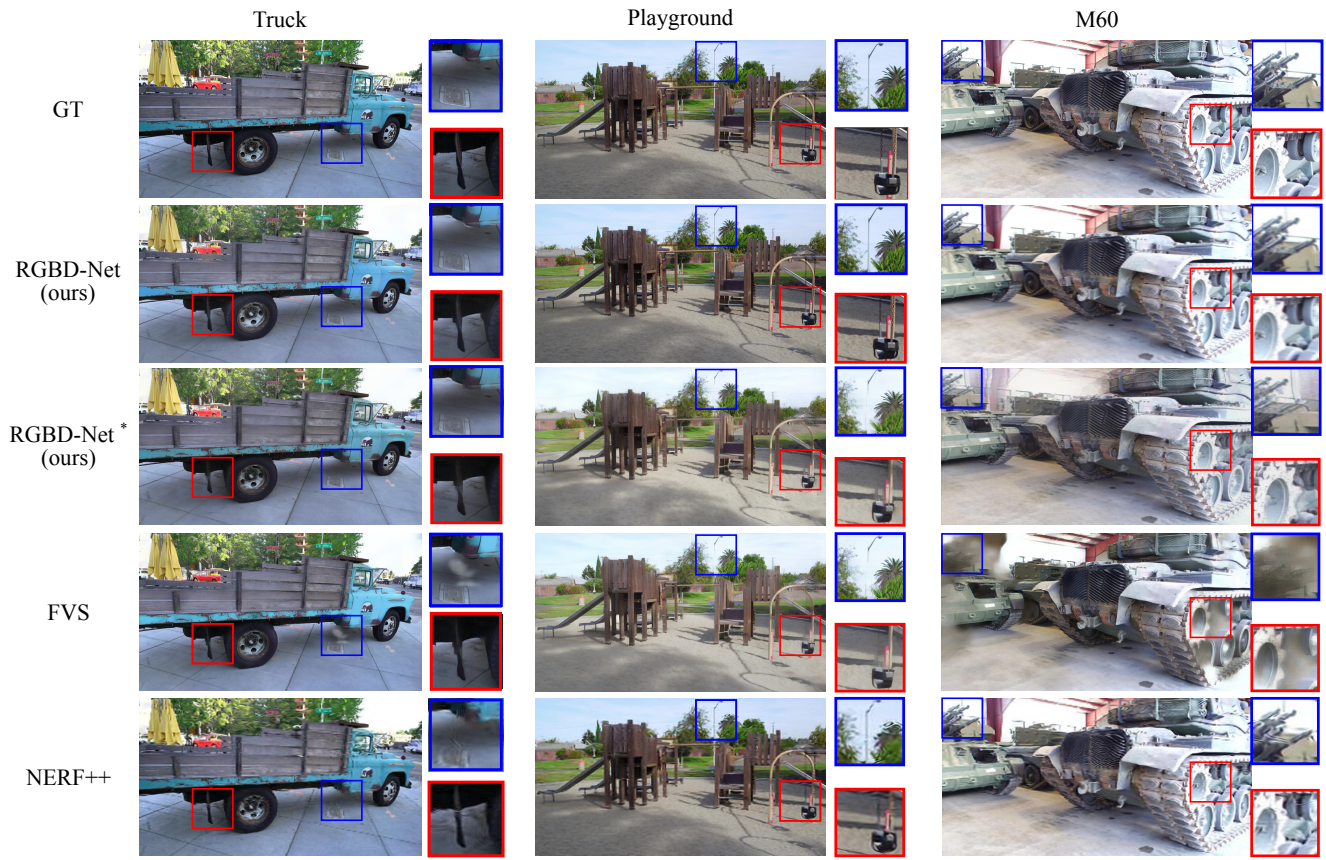}
\end{center}
  \caption{Additional qualitative results on Tanks and Temples dataset \cite{TankAndTemples}. RGBD-Net$^*$ is our proposed RGBD-Net trained without the ground-truth depth loss. We observer no significant difference on the performance of view synthesis between RGBD-Net and RGBD-Net$^*$.}
\label{fig_qualTT}
\end{figure*}
\begin{figure*}[t]
\begin{center}
  \includegraphics[width=\linewidth]{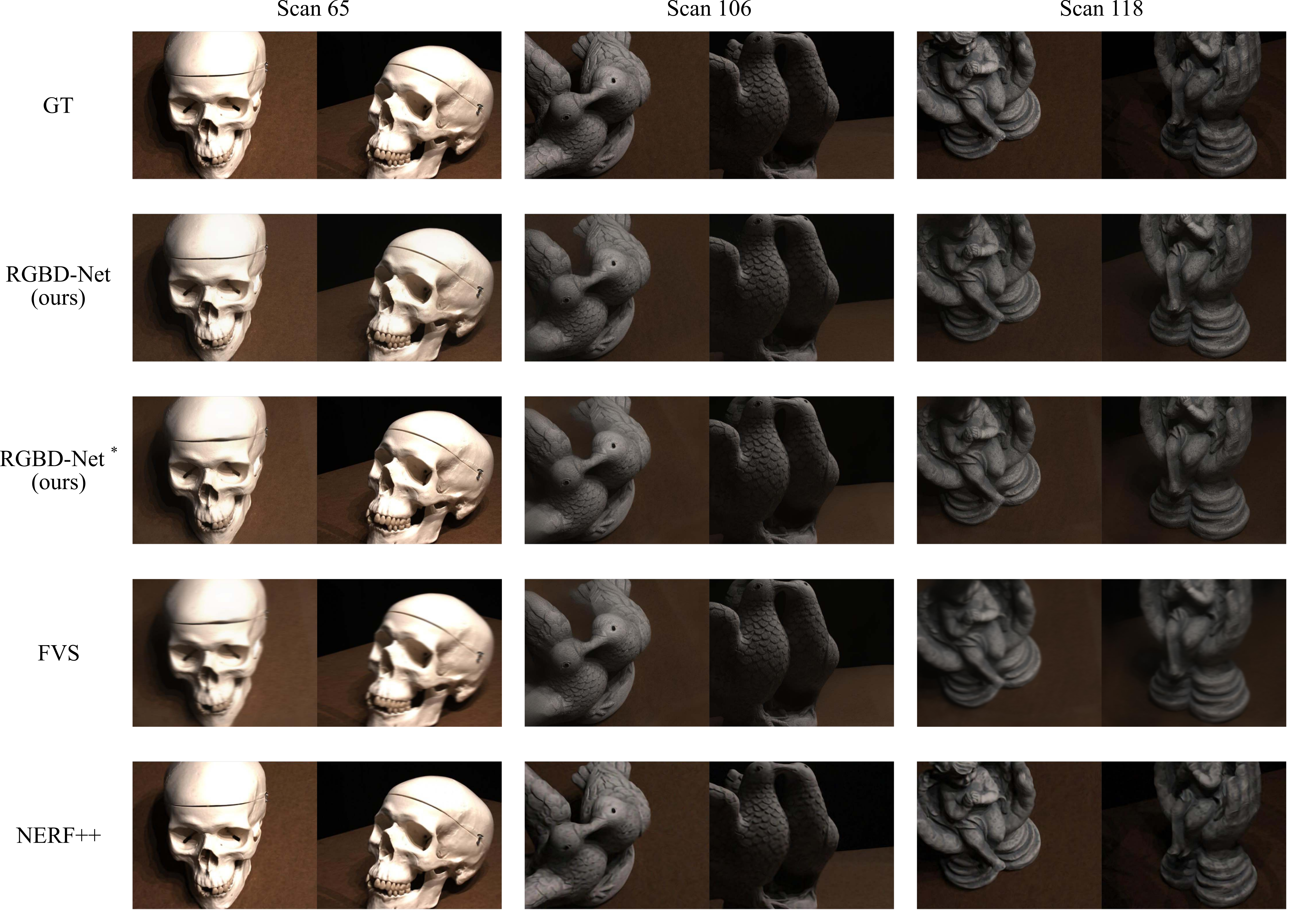}
\end{center}
  \caption{Additional qualitative results on DTU dataset \cite{DTU}.}
\label{fig_qualDTU}
\end{figure*}
\begin{figure*}[t]
\begin{center}
  \includegraphics[width=\linewidth]{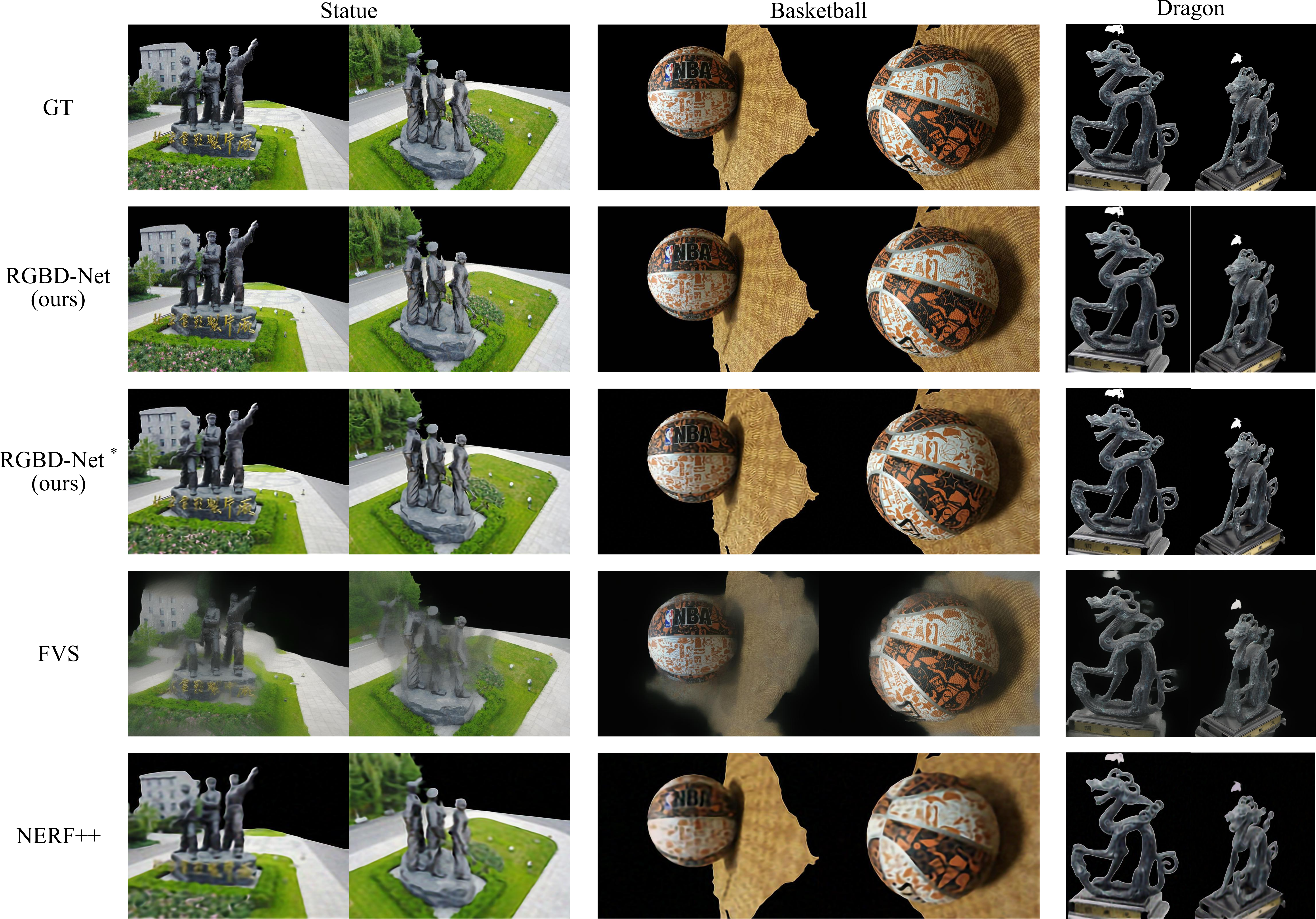}
\end{center}
  \caption{Additional qualitative results on BlendedMVS dataset \cite{yao2020blendedmvs}.}
\label{fig_qualBlend}
\end{figure*}

In this section, we provide additional qualitative results. Fig. \ref{fig_qualTT}, \ref{fig_qualDTU} and \ref{fig_qualBlend} show more examples of rendered novel views using RGBD-Net and other view synthesis methods on the Tanks and Temples \cite{TankAndTemples}, DTU \cite{DTU} and BlendedMVS \cite{yao2020blendedmvs} datasets, respectively. For NeRF~\cite{Nerf}, we manually define the bounding volume around the main object in each testing scene.

\end{document}